%% file: ms.tex
\documentclass[11pt,a4paper,reqno]{amsart}
\usepackage{amsmath,amssymb,amsbsy,amsthm,dsfont}
\usepackage{graphicx,a4wide,hyperref}
\usepackage{multirow}
\usepackage[utf8]{inputenc}
\usepackage[T1]{fontenc}
\usepackage{xcolor}
\usepackage{float}
\usepackage{algorithm,algorithmic}

\usepackage{url}

\newcounter{hypA}

\def\rset{\mathbb{r}}

\def\argmax{\mathop{\rm Arg\max}\limits}

\def\rset{\mathbb{R}}

\def\nset{\mathbb{N}}
\def \1{\mathbf{1}}

\begin{document}

\title[An active learning approach for equilibrium based chemical simulations]{An active learning approach for improving the performance of equilibrium based chemical simulations}
\author{Mary Savino}
\address{Andra, 1/7 Rue Jean Monnet, 92290 Châtenay-Malabry, France and Université Paris-Saclay, AgroParisTech, INRAE, UMR MIA-Paris,
  75005, Paris, France}
\author{Céline Lévy-Leduc}
\address{Université Paris-Saclay, AgroParisTech, INRAE, UMR MIA-Paris,
  75005, Paris, France}
\email[Corresponding author]{celine.levy-leduc@agroparistech.fr}
\author{Marc Leconte}
\address{Andra, 1/7 Rue Jean Monnet, 92290 Châtenay-Malabry, France}
\author{Benoit Cochepin}
\address{Andra, 1/7 Rue Jean Monnet, 92290 Châtenay-Malabry, France}

\begin{abstract}

\input{abstract_revision.tex}

\end{abstract}


\maketitle

\section{Introduction}

\input{intro_revision.tex}

\section{Description of our approach}\label{sec:method}

\input{method_revision.tex}

\section{Numerical experiments}\label{sec:numexp}

\input{numexp_revision.tex}

\section{Application to a  multidimensional geochemical system}\label{sec:real}

\input{real_revision.tex}

\section{Conclusion}

\input{conclusion_revision.tex}

\newpage

\section{Appendix: Additional plots}

\input{appendix_revision.tex}

\end{document}

%% file: abstract_revision.tex
In this paper, we propose a novel sequential data-driven method for dealing with equilibrium based chemical simulations, which can be seen as a specific machine
learning approach called active learning. The underlying idea of our approach is to consider the function to estimate as a sample of a Gaussian process which allows us
to compute the global uncertainty on the function estimation. Thanks to this estimation and with almost no parameter to tune, the proposed method sequentially
chooses the most relevant input data at which the function to estimate has to be evaluated to build a surrogate model. Hence, the number of evaluations of the function to estimate is dramatically limited.
Our active learning method is validated through numerical experiments and applied to a complex chemical system commonly used in geoscience.


%% file: intro_revision.tex
Computing the concentrations at equilibrium of reactive species is well known to be a challenging issue when the number
of species is high and/or when the reaction involves the dissolution or the precipitation of minerals \cite{white58,smith80,decapitani87}.
The numerical resolution of these non-linear problems can quickly become so time consuming
that the coupling with other physical processes has to be simplified.
For instance in the case of reactive transport, it means that the size of the geometric model has to be drastically
limited leading typically to a one dimensional model or that the number of time steps has to be reduced.
To overcome this issue, research efforts have been dedicated to the improvement of the
numerical scheme aiming at speeding up the computations. A classical approach consists in using a splitting operator technique
to solve separately the transport of the chemical species and the chemical reaction between those species
\cite{marchuk90,sportisse00,descombes01,carrayrou04,simpson07}. With this approach a specific optimization for each part of the resolution can be performed
especially by taking advantage of the parallel architecture of computers \cite{farago07,geiser11,geiser20}.

However, despite the significant improvements of the numerical solvers and preconitionners during the last decades,
three dimensional large scale modelling of complex reactive transport over a long period of time, namely many time steps,
remains almost impossible to solve with standard computers.
Consequently, the recent success of machine learning (ML) in various fields have quickly drawn attention of geoscientists because ML seems to be able
to solve very complex problems with a reasonable cost in terms of computational ressources.

The main idea behind the ML success is to provide an estimation of the solution of the full simulation model
that can replace it.
Two of the most popular approaches are model order reduction and data-driven models also called surrogate models.
The first one requires to understand the underlying
chemical processes to create a simplified model while preserving some physical principles \cite{rao13}.
In the second approach, the underlying chemical processes are not assumed to be known or understood and a model is solely built
from a limited but potentially significant set of values of the solution of the full simulation model associated to some specific input values  \cite{guerillot20}.
Since the number of required values is unknown a priori,
choosing the optimal input values and parameters used for building the surrogate model is crucial and usually challenging.

In this paper, we propose a novel sequential data-driven method for dealing with equilibrium based chemical simulation, which can thus be seen as an active
learning approach inspired by the ideas contained in \cite{srinivas:2012,jala:2016}. With such an approach, our goal is
to minimize the number of evaluations of the function that has to be estimated to build a surrogate model.
Our approach consists in modeling the function to estimate as a sample of a Gaussian Process (GP) which allows us to provide an error estimation to
sequentially choose the most relevant input data until a given stopping criterion is fulfilled. The advantage of our approach is that the number of required
evaluations of the function to estimate is very limited and that there are no parameter to tune.

The paper is organized as follows. In Section \ref{sec:method}, our approach is described. Some numerical experiments are provided in Section \ref{sec:numexp}
to illustrate the statistical and numerical performance of our method. It is then applied in Section \ref{sec:real} to a multidimensional example coming from \cite{kolditz12} which includes several chemical elements and minerals.

%% file: method_revision.tex
In this section, we describe our active learning approach for estimating
a real-valued function $f$ defined on a compact subset $\mathcal{A}\subset\rset^d$ by using
only a few number of sequentially well-chosen points at which $f$ is evaluated.

We adopt a Bayesian point of view which consists in
considering $f$ as a sample of a zero-mean Gaussian process (GP)
having a covariance function $k$ that
we shall denote by GP(0,$k(\cdot,\cdot)$) in the following.
The advantage of this approach is that, conditionally
on a set of $t$ observations $\mathbf{y}_t=(y_1,\dots,y_t)'$ where
$y_i=f(x_i)$, $x_i$ belonging to $\mathcal{A}$, the posterior distribution is still a GP
having a mean $\mu_t$ and a covariance function $k_t$ given by
\begin{align}\label{eq:muT}
\mu_t(u)&=\mathbf{k}_t(u)'\mathbf{K}_t^{-1}\mathbf{y}_t\;,\\
\label{eq:kT}
k_t(u,v)&=k(u,v)-\mathbf{k}_t(u)'\mathbf{K}_t^{-1}\mathbf{k}_t(v)\;,
\end{align}
where $\mathbf{k}_t(u)=[k(x_1,u) \dots k(x_t,u)]'$. Here $'$ denotes the matrix transposition, $u$ and $v$ are in
$\mathcal{A}$ and
$\mathbf{K}_t=[k(x_i,x_j)]_{1\leq i,j \leq t}$, where the $x_i$'s are in $\mathcal{A}$.
For further details on GP, we refer the reader to \cite{rasmussen:2006} in which their properties are thoroughly presented.

In our case, $f$ models a physical quantity that is assumed to be smooth, so for our applications we shall consider two covariance functions that are commonly used in this case. The first one is the squared exponential (SE) covariance function
\begin{equation}\label{eq:cov_gauss}
k_{\mathrm{SE}}(u,v)=\exp\left(-\frac{1}{2}(u-v)'M^{-1}(u-v)\right)\;,
u,v\in\mathcal{A}\subset\rset^d\;,
\end{equation}
\begin{equation}\label{eq:matlengthscale}
M=diag\left(\ell_1^2,\dots,\ell_d^2\right)\;,\;\ell_1\;,\;\ell_2\;,\ldots\;,\ell_d>0\;.
\end{equation}

Here the $\ell_1\;,\;\ell_2\;,\ldots\;,\ell_d$ hyperparameters are the characteristic length scales.
Actually, these hyperparameters can be understood as how far you need to move along a particular axis in the
  input space so that the function values become uncorrelated. For further details, we refer the reader to Section 5.1 of
\cite{rasmussen:2006}.
Note that Definition~\eqref{eq:cov_gauss} allows us to model anisotropic response surfaces.

As explained in  \cite{rasmussen:2006}, since this covariance function is infinitely differentiable, the GP with this
covariance function has mean square derivatives of all orders. As argued by
\cite{Stein99}
 such strong smoothness
assumptions may be unrealistic for modeling many physical processes, so we shall
also consider another covariance function belonging to the Mat\' ern class of covariance functions
defined by
\begin{equation}\label{eq:cov_matern}
k_{\textrm{Mat\' ern}}(r)=\frac{2^{1-\nu}}{\Gamma(\nu)}\left(\sqrt{2\nu}r\right)^{\nu}K_{\nu}\left(\sqrt{2\nu}r\right)\;, \nu>0\;,
\end{equation}
where $K_{\nu}$ is a modified Bessel function with Bessel order $\nu$, see \cite[Section 9.6]{Abramowitz65}, and $r$ is defined by
\begin{equation}
\label{eq:rmatern}
r=\sqrt{(u-v)'M^{-1}(u-v)}\;,
u,v\in\mathcal{A}\;,
\end{equation}
$M$ being defined in \eqref{eq:matlengthscale}.
In this situation, as explained in \cite{rasmussen:2006}, the GP is $q$-times mean-square differentiable if and only if $\nu > q$.
Here, we shall focus on the case
where $\nu=5/2$, for which $k_{\textrm{Mat\' ern}}$ has a computationally advantageous expression. Indeed, for $\nu=p+\frac{1}{2}$, where
$p$ is in $\nset$,
\begin{equation}\label{eq:cov_maternp}
k_{\textrm{Mat\' ern}}(r)=\exp\left(-\sqrt{2\nu}r\right)\frac{\Gamma(p+1)}{\Gamma(2p+1)}\sum^p_{i=0}\frac{(p+i)!}{i!(p-i)!}\left(\sqrt{8\nu}r\right)^{p-i}\;,
\end{equation}
with $r$ defined in \eqref{eq:rmatern}; see \cite[Equation 10.2.15]{Abramowitz65} for further details.

In the following, we shall denote by $\textrm{A}$ a fine grid of $\mathcal{A}$:
\begin{equation}\label{eq:def_A}
\mathrm{A}= \{\mathrm{x}_1,\dots,\mathrm{x}_m\} \subset \mathcal{A}\;.
\end{equation}
This grid is either a regular grid of $\mathcal{A}\subset\mathbb{R}^{d}$ when $d$ is small (usually 1 or 2) or a
  Latin Hypercube Sampling for larger values of $d$. Note that this grid contains the points at which the estimation of $f$ is performed and
that the points at which $f$ is evaluated are chosen in this grid.

Inspired by \cite{srinivas:2012} who proposed a sequential approach for maximizing a function by modeling it using a Gaussian process, we propose
a strategy which consists in adding the new point $x_{t+1}$ to the set of $t$ observations at which $f$ needs to be evaluated as follows:
\begin{equation}\label{eq:pureexploration}
x_{t+1}\in\argmax_{x\in\mathrm{A}}\sigma_t(x)\;,
\end{equation}
where
\begin{equation}\label{eq:sigma_t}
\sigma_t(x)^2=k_t(x,x), 
\end{equation}
$k_t$ being defined in~(\ref{eq:kT}) and $\argmax_{x\in\mathrm{A}}\sigma_t(x)$ being the set of $x\in\mathrm{A}$ where $\sigma_t(x)$ reaches its maximum.
Note that the points $x_1,x_2,\dots,x_t,x_{t+1},\dots$ at which $f$ needs to be evaluated are chosen in the fine grid $\mathrm{A}$ of $\mathcal{A}$
defined in (\ref{eq:def_A}).

\subsection{Estimating the characteristic length scales}\label{subsec:estl}

Previously, we assumed that the characteristic length scales $\boldsymbol{\ell}=(\ell_i)_{\{1\leq i\leq d\}}$ were known.
However, this is obviously not the case in real-data applications. We propose using the maximum-likelihood strategy described in 
\cite{rasmussen:2006} to estimate $\boldsymbol{\ell}$.
This adds a step to the method previously described, as the $\ell_i$'s have to be estimated before evaluating the 
posterior distribution of the GP using (\ref{eq:muT}) and (\ref{eq:kT}).
Hence, for the observation set $\{(x_1,y_1),\ldots,(x_t,y_t)\}$ with $y_i=f(x_i)\;,1\leq i\leq t$, the posterior log-likelihood 
given by:
\begin{equation}\label{eq:log_lik}
-\frac{1}{2}\mathbf{y}'_t\mathbf{K}_t^{-1}\mathbf{y}_t-\frac{1}{2}\log|\mathbf{K}_t|-\frac{t}{2}\log2\pi\;,
\end{equation}
with $\mathbf{y}_t=(y_1,\ldots, y_t)'$ and $\mathbf{K}_t=[k(x_i,x_j)]_{1\leq i,j \leq t}$, has to be maximized with respect to $\boldsymbol{\ell}$.

\subsection{Summary of our strategy}


Our method was implemented by using the GaussianProcessRegressor class of the \texttt{scikit-learn 0.20.3}
module of Python which only provides the computation of $\mu_t$ and $\sigma_t$ defined in (\ref{eq:muT}) and (\ref{eq:sigma_t}).
Our sequential approach is summarized in Algorithm \ref{algo:gps}.

\begin{algorithm}
\caption{}
\label{algo:gps}
\begin{flushleft}
\textbf{Input:} $x_1,\ldots,x_{t_1}$ a small initial set of points of $\mathrm{A}$ where $f$ has been evaluated 
\\
$t=t_1$; Choose a covariance function $k$ among SE and Mat\'ern.\\
\textbf{While} the stopping criterion is not fulfilled 
\begin{itemize}
\item Estimate $\boldsymbol{\ell}$ by using (\ref{eq:log_lik})
\item Evaluate the posterior distribution of the GP using (\ref{eq:muT}) and (\ref{eq:kT}), and the variance $\sigma_{t}(x)^2$ for all $x$ in
  $\textrm{A}$
\item Choose $x_{t+1}$ in $\textrm{A}$ using (\ref{eq:pureexploration})
\item Evaluate $f$ at this point: $y_{t+1}=f(x_{t+1})$
\item Add this new observation to the set of points at which $f$ is evaluated which becomes $x_1,\ldots,x_{t},x_{t+1}$
\item $t\leftarrow t+1$
\end{itemize}
The function $f$ is estimated by $\mu_t$ defined in (\ref{eq:muT}).
\end{flushleft}
\end{algorithm}

Further comments on the stopping criteria appearing in Algorithm \ref{algo:gps} are given below.

\subsection{Stopping criteria}\label{subsec:stop_crit}

Different stopping criteria based on the following quantities can be used.

\begin{itemize}
\item \textbf{\textsf{Ratio variance.}}
  At each iteration $t$ of our method, the following average is computed:
  \begin{equation}\label{eq:Rk}
  R_n(t)=\frac{1}{n-1}\sum_{i=1}^{n-1}\frac{\max_{x\in\textrm{A}}\sigma^{2}_{t}(x)}{\max_{x\in\textrm{A}}\sigma^{2}_{t-i}(x)},
 \end{equation}
where $\sigma_t$ is defined in (\ref{eq:sigma_t}) and $n=2$, $5$ or $10$.  This criterion will be then compared to a threshold to determine if the maximal variance reach a plateau. In some cases, $\sigma^2_{t-i}$ can be less than $\sigma^2_{t}$ so in order to detect the smallest variations, we also have to make sure that the ratio does not exceed the inverse of the chosen threshold. Thus, the associated stopping criterion is: interrupt the algorithm when $t$ is such that
  \begin{equation}\label{eq:Rk_seuil}
  0.9<R_n(t)<\frac{1}{0.9}.
\end{equation} 
\item \textbf{\textsf{Mobile average.}}
  At each iteration $t$ of our method, the following average is computed:
  \begin{equation}\label{eq:Ml}
  M_\ell(t)=\frac{1}{\ell}\sum_{j=0}^{\ell-1}\max_{x\in\textrm{A}}\sigma^2_{t-j}(x)
\end{equation}
for $\ell=5$ or $10$ where $\sigma_t$ is defined in (\ref{eq:sigma_t}).
 The associated stopping criterion is: interrupt the algorithm when $t$ is such that
  \begin{equation}\label{eq:Ml_seuil}
  M_\ell(t)<0.01.
\end{equation}
\item \textbf{\textsf{Maximal variance.}}
  At each iteration $t$ of our method,
  \begin{equation}\label{eq:V}
  V(t)=\max_{x\in\textrm{A}}\sigma^2_{t}(x)
\end{equation}
is computed where $\sigma_t$ is defined in (\ref{eq:sigma_t}). The associated stopping criterion is:
  interrupt the algorithm when $t$ is such that
  \begin{equation}\label{eq:V_seuil}
  V(t) < s,
\end{equation}
where $s=0.01$ or 0.001 in the following.
\end{itemize}

The statistical performance of these different criteria are investigated in Section \ref{sec:numexp}.
Note that the values reported here for each criteria ($0.9$, $0.01$ or $0.001$) were
chosen based on some numerical experiments since they appear to be relevant to detect a plateau in the maximal variance.


%% file: numexp_revision.tex
To illustrate our method we consider hereafter the estimation of the amount of a "Salt" mineral as a function of the concentrations of its constituents Sp$_{a}^{+}$ and Sp$_{b}^{-}$. For this example, the thermodynamic constants of the halite salt (NaCl) were considered because there are only two constitutive
elements and because they do not depend on the pH of the solution. From our point of view, there is no theoretical limitation in the application
of our method to more complex salts or minerals.

\begin{figure}
\begin{center}
  \includegraphics[scale=0.45]{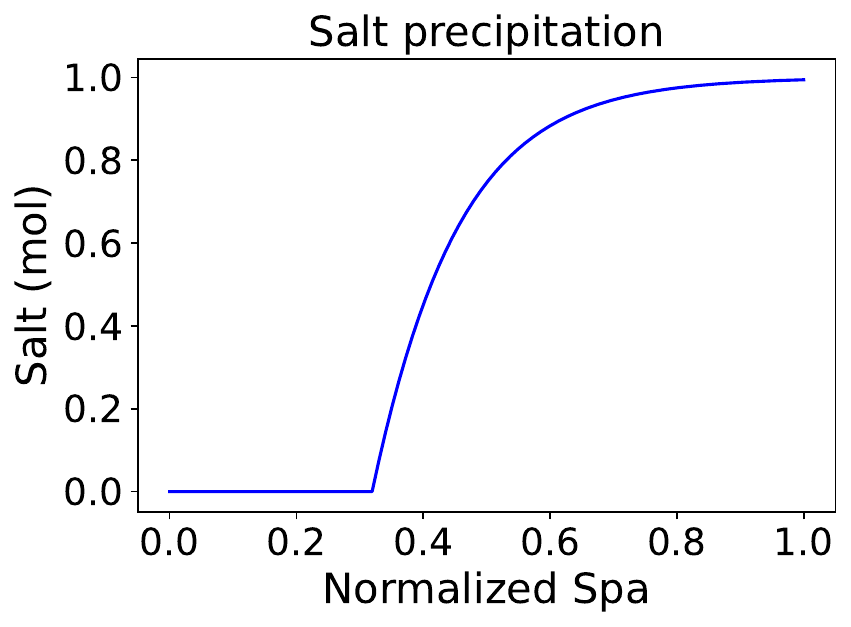}
  \includegraphics[scale=0.26,clip, trim = {0 0 0 4cm}]{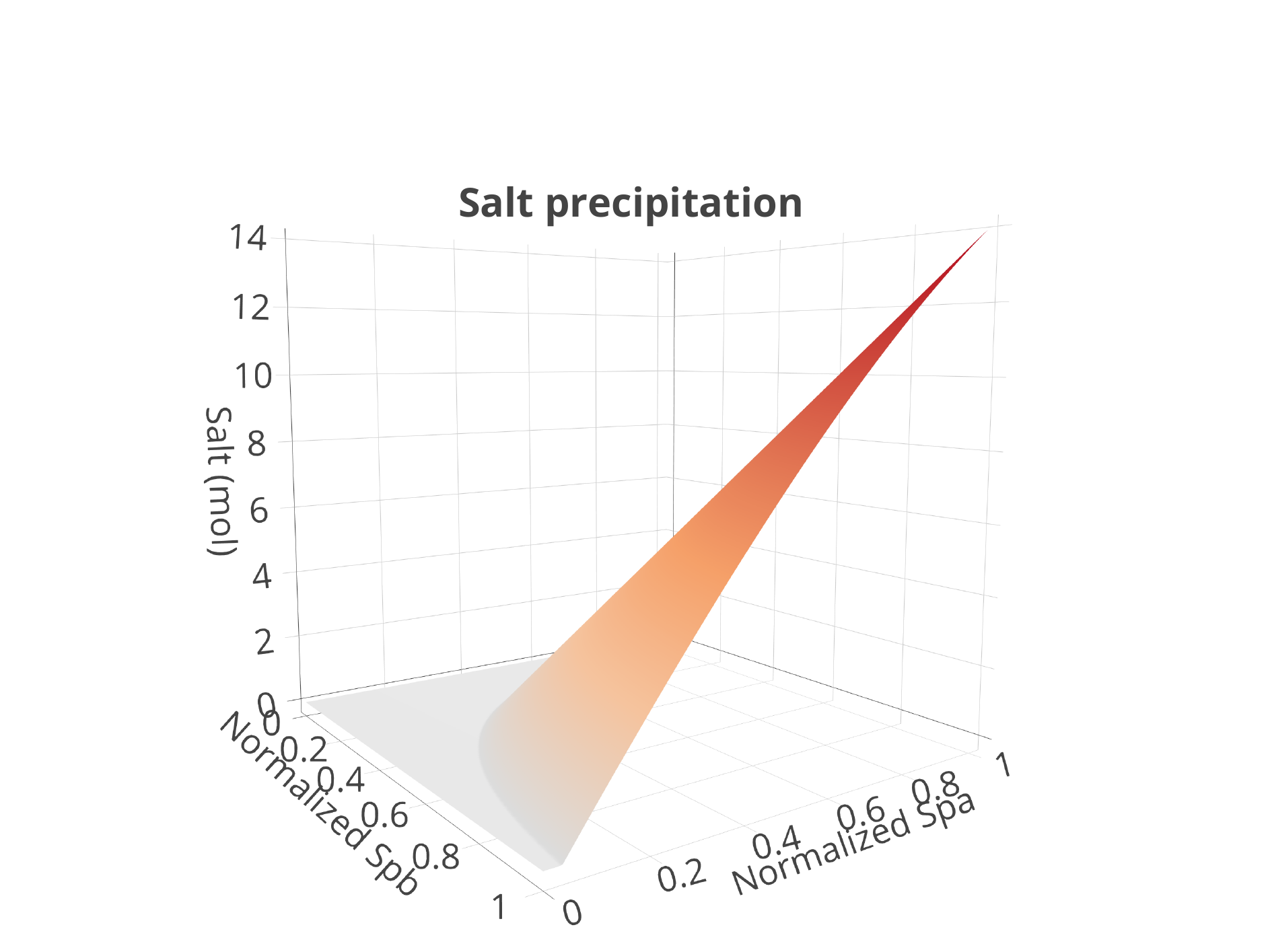}
\caption{Functions $f$ to estimate when $d=1$ (left) and $d=2$ (right). \label{fig:courbes}}
\end{center}
\end{figure}

Following the law of mass action, the dissolution reaction of this mineral writes:

\[\mathrm{
   Salt \rightleftharpoons Sp_{a}^{+} + Sp_{b}^{-}.
}\]
At equilibrium, the activity of these elements a$_{Sp_{a}^{+}}$ and a$_{Sp_{b}^{-}}$ obey the solubility product
\[\mathrm{
   K_{Salt} = a_{Sp_{a}^{+}} a_{Sp_{b}^{-}} = 10^{1.570}.
}\]
The amount of Salt was first calculated with PHREEQC \cite{Parkhurst2013} as a function of the concentrations of Sp$_{a}^{+}$, which is normalized so that $\mathcal{A}=[0,1]$. It corresponds to the case $d=1$ below.
The corresponding function $f$ is displayed in the left part of Figure \ref{fig:courbes} where $\mathrm{A}$ is a regular grid of $\mathcal{A}$
with $m = 1140$ points. Then, the amount of Salt was computed with PHREEQC
  as a function of the concentrations of Sp$_{a}^{+}$ and Sp$_{b}^{-}$, which  are also normalized so that $\mathcal{A}=[0,1]^2$. It corresponds to the case $d=2$ below. The corresponding function $f$ is displayed in the right part of Figure
  \ref{fig:courbes} where $\mathrm{A}$ is a regular grid of $\mathcal{A}$
with $m = 40000$ points.

\subsection{Case $d=1$}

\begin{figure}
\begin{center}
\includegraphics[scale=0.3]{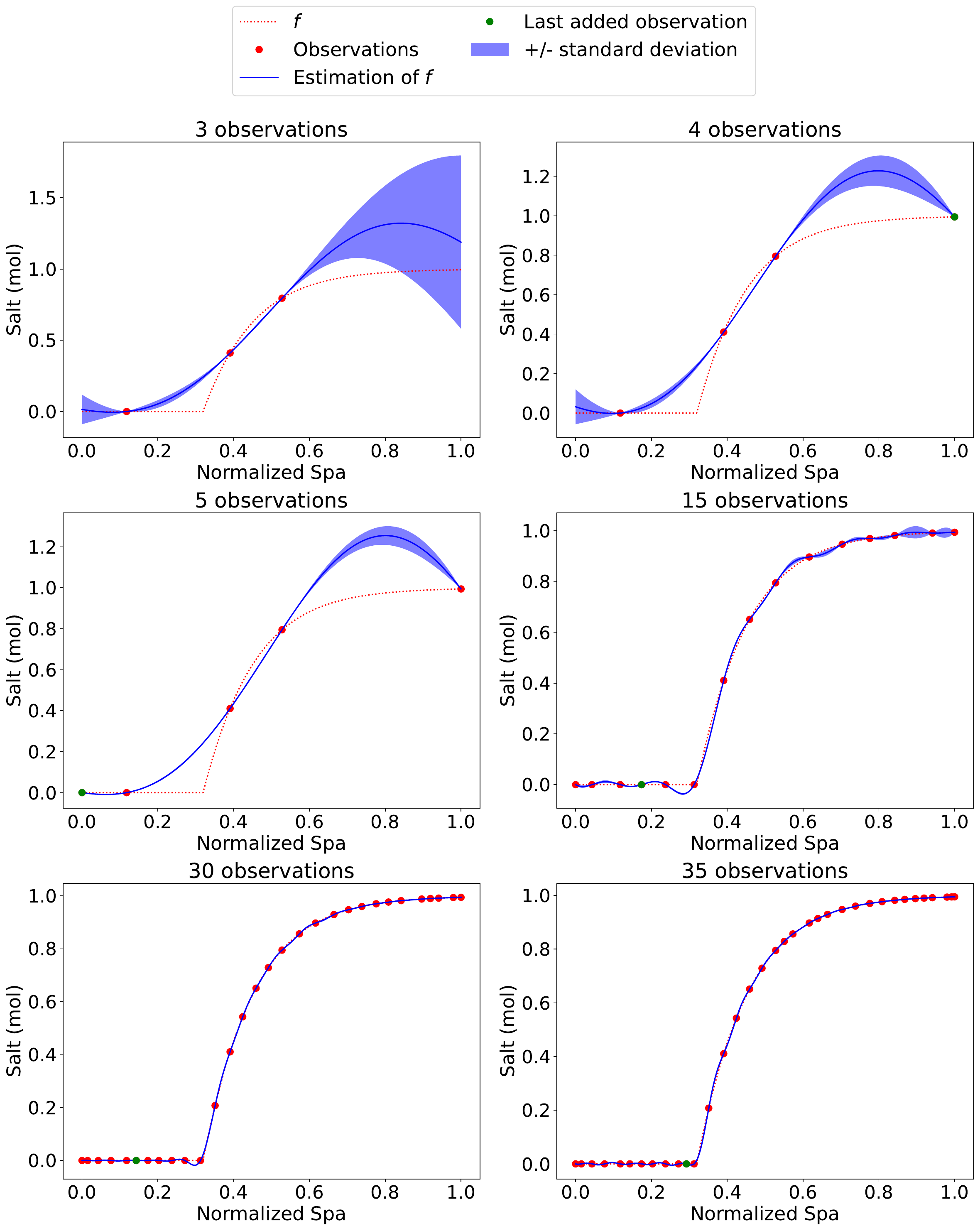}
\caption{Illustration of our active learning approach for estimating the function displayed in the left part of Figure \ref{fig:courbes} by starting
  from $t_1=3$ observations randomly chosen in $\textrm{A}$ with the squared exponential covariance function.} \label{fig:illustration}
\end{center}
\end{figure}

The different steps of our approach summarized in Algorithm \ref{algo:gps} are illustrated in Figure \ref{fig:illustration}
where our procedure was arbitrarily stopped after $40$ evaluations.
Here, we used the SE covariance function defined in (\ref{eq:cov_gauss}).

The approach starts with $t_1=3$ points randomly chosen in $\mathrm{A}$.
Then, a new point in green is added to the set of points at which an evaluation of $f$ is required. This point corresponds
to the position on the $x$-axis where the uncertainty $\sigma_{t}^2$ associated to the estimation of $f$ is maximized.
We can see from this figure which displays the true function $f$, the estimation of $f$ and the points at which $f$ has been evaluated that
$35$ evaluation points are enough to obtain a very accurate estimation of $f$.

To further investigate the statistical performance of our approach, we used the following measures:

 \begin{equation}\label{eq:MAE_norm}
  \textrm{Normalized MAE}(t)=\frac{1}{m}\sum_{i=1}^m \frac{\left|y_i-\mu_{t}(x_i)\right|}{y_{max} - y_{min}}, 
\end{equation}
where $\mu_t$ is the estimation of $f$ obtained at iteration $t$, $m$ is the number of elements in the grid $\textrm{A}$ and $y_{min}$ and $y_{max}$ are the minimum and maximum values, respectively, found for the evaluation of $f$ on the initial grid ;



\begin{equation}\label{eq:norm_sup_norm}
  \textrm{Normalized sup norm}(t)=\max_{1\leq i\leq m} {\frac{\left|y_i-\mu_{t}(x_i)\right|}{y_{max} - y_{min}}}.
\end{equation}

\begin{equation}\label{eq:V}
V(t)=\max_{x\in\textrm{A}}\sigma^2_{t}(x),
\end{equation}
where $\sigma_t$ is defined in (\ref{eq:sigma_t}).

\begin{figure}
\begin{center}
  \includegraphics[scale=0.35,trim={0 0 0 1.5cm},clip]{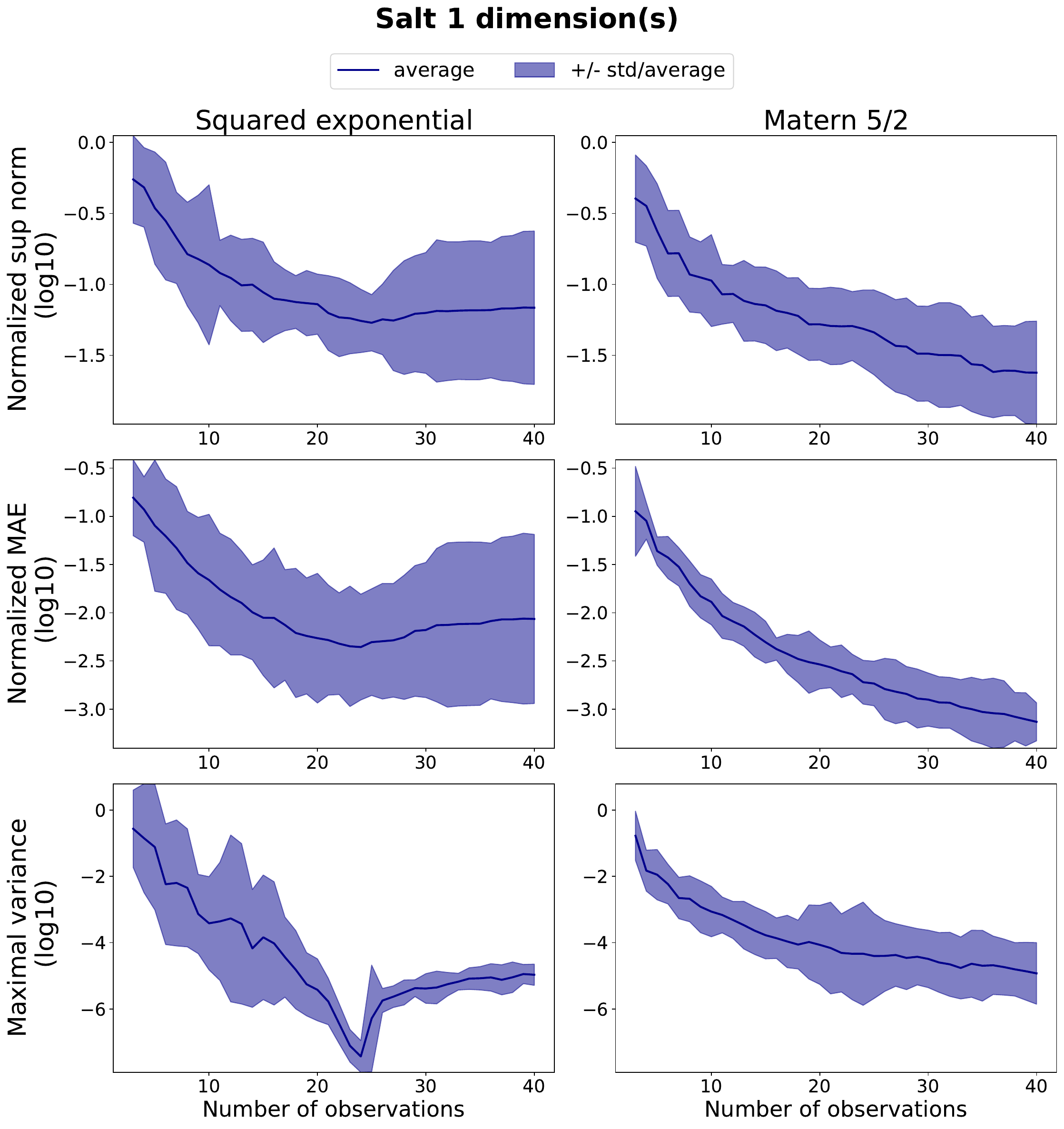}
  \caption{Average and standard deviation of different statistical measures for the squared exponential covariance function defined in (\ref{eq:cov_gauss}) (left)
      and for  the Matern covariance function defined in (\ref{eq:cov_maternp}) (right) in the case $d=1$. \label{fig:crit1D}}
\end{center}
\end{figure}

The average and the standard deviation of these measures obtained from 10 replications of the initial set of points are displayed in 
Figure \ref{fig:crit1D} for the covariance functions defined in (\ref{eq:cov_gauss}) and (\ref{eq:cov_maternp}) and $3\leq t\leq 40$.
Note that the average and the standard deviation are computed by using 10 different initial sets of points.

We can see from this figure that the performance of our approach is slightly better for the Mat\'ern covariance function  than for the squared
exponential function.  It can indeed reach a normalized MAE  (resp. normalized sup norm) of $10^{-3}$ (resp. $10^{-1.5}$) by using only 40 evaluations of the function
to estimate. This might come from the discontinuity of the first derivative of the function to estimate where the salt starts to precipitate.

\begin{figure}
\begin{center}
\includegraphics[scale=0.4]{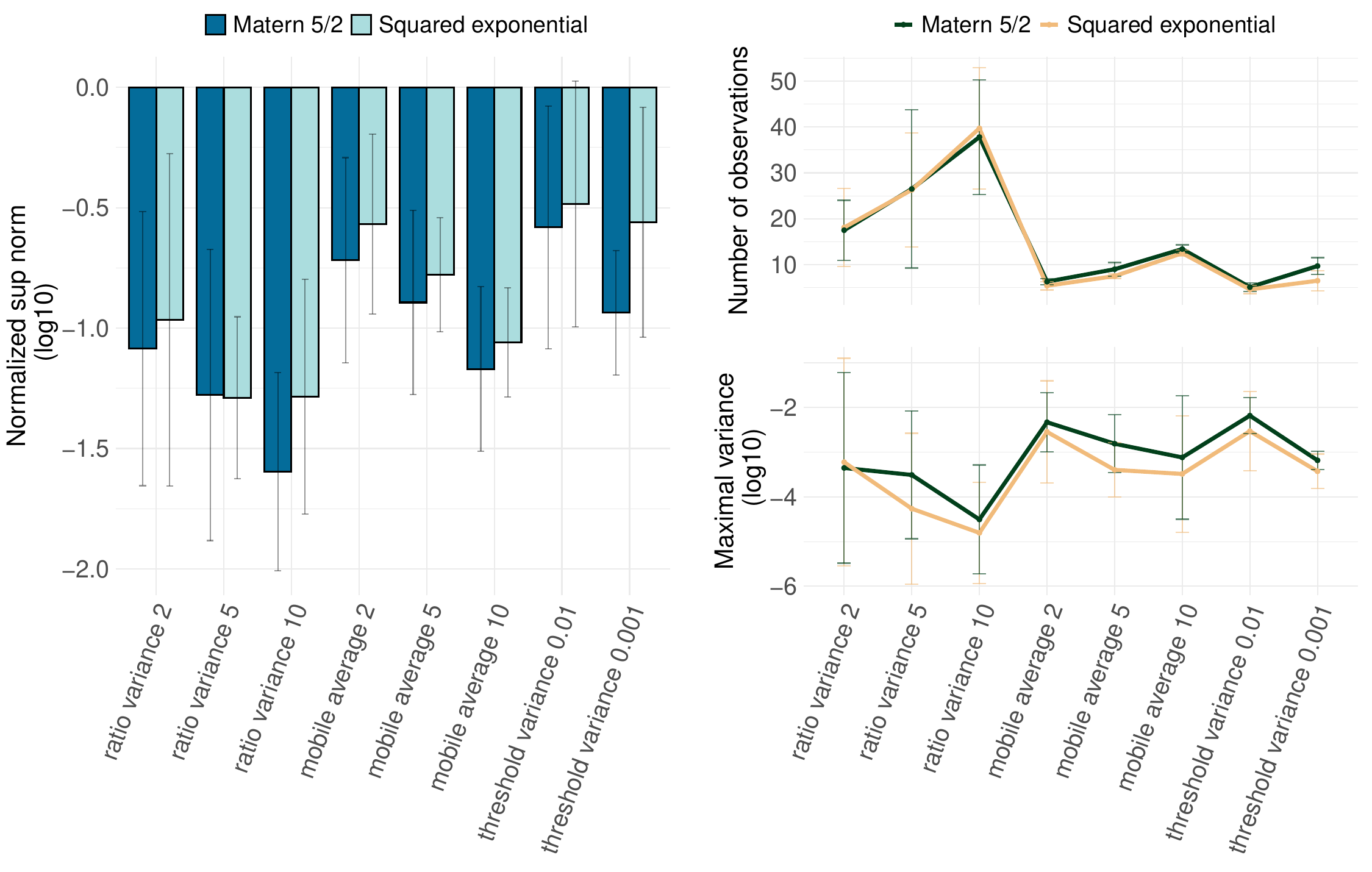}
\caption{Left: Statistical assessment of the error estimation of $f$ displayed in the left part of Figure \ref{fig:courbes} for the stopping criteria
  defined in (\ref{eq:Rk_seuil}), (\ref{eq:Ml_seuil}) and (\ref{eq:V_seuil}) for the squared exponential and the Mat\'ern covariance functions.
  Top right: Number of evaluations required for the considered stopping criteria. Bottom right: Values of $V(t^\star)$
  where $V$ is defined in (\ref{eq:V}) and $t^\star$ is the stopping iteration which changes from one stopping criterion to another. \label{fig:crit_sum_1D}}
\end{center}
\end{figure}

In the left part of Figure \ref{fig:crit_sum_1D} the statistical performance of our approach including the stopping criteria
are further investigated thanks to the computation
of the previous performance measures defined in (\ref{eq:MAE_norm}), (\ref{eq:norm_sup_norm}) and (\ref{eq:V}):
$\textrm{Normalized MAE}(t^\star)$, $\textrm{Normalized Sup norm}(t^\star)$ and $V(t^\star)$ where $t^\star$ is the stopping iteration
which may be different for each stopping criterion.

We can see from the left part of Figure \ref{fig:crit_sum_1D} that among all of the stopping criteria,
``ratio variance 5'' ($R_5$), ``ratio variance 10'' ($R_{10}$) and ``mobile average 10'' ($M_{10}$) are those providing the best estimations of the function $f$.
Moreover, we can observe from the right part of this figure that our active learning approach
only requires between 15 and 40 evaluations of the function to estimate 
instead of the 1140 points of the initial grid to provide a very accurate
estimation of the function $f$. With such an approach, we can thus expect a significant reduction of the computational time especially in situations where
the computational load associated to the evaluation of $f$ is high.
Figure \ref{fig:crit_sum_1D} also shows that, in this case, the impact of the covariance function is not significant
  even though the first derivative of the function to approximate is not continuous, namely where the salt precipitates.

\subsection{Case $d=2$} \label{subsec:halite_2D} 
In order to further assess the performance of our approach we now consider the estimation of
the amount of Salt as a function of the concentrations of Sp$_{a}^{+}$ and Sp$_{b}^{-}$.

\begin{figure}
\begin{center}
 \includegraphics[clip, scale=0.3, trim=4.5cm 0.2cm 5cm 3cm]{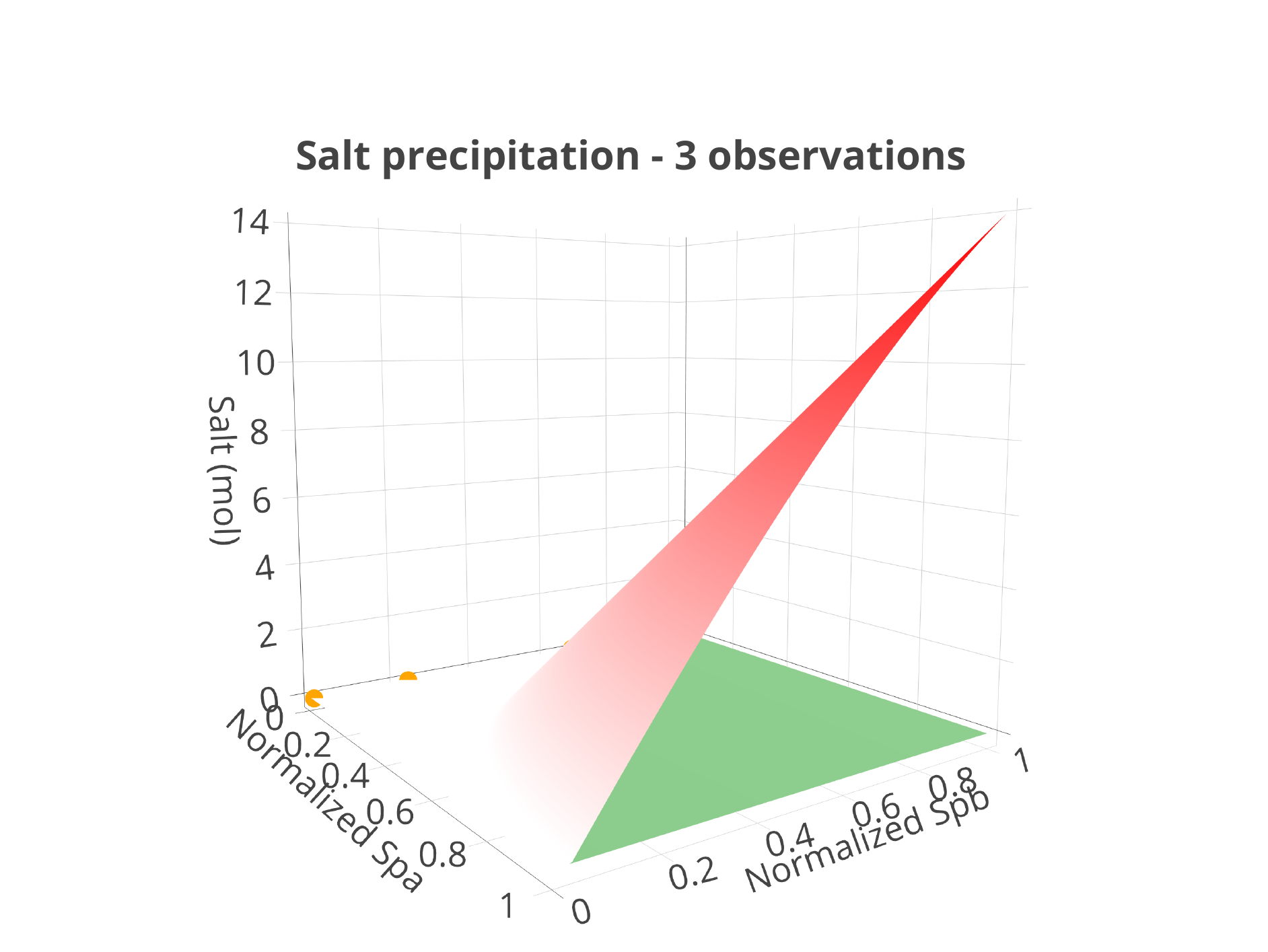} \hspace{1em}
  \includegraphics[clip, scale=0.3, trim=4cm 0.2cm 5cm 3cm]{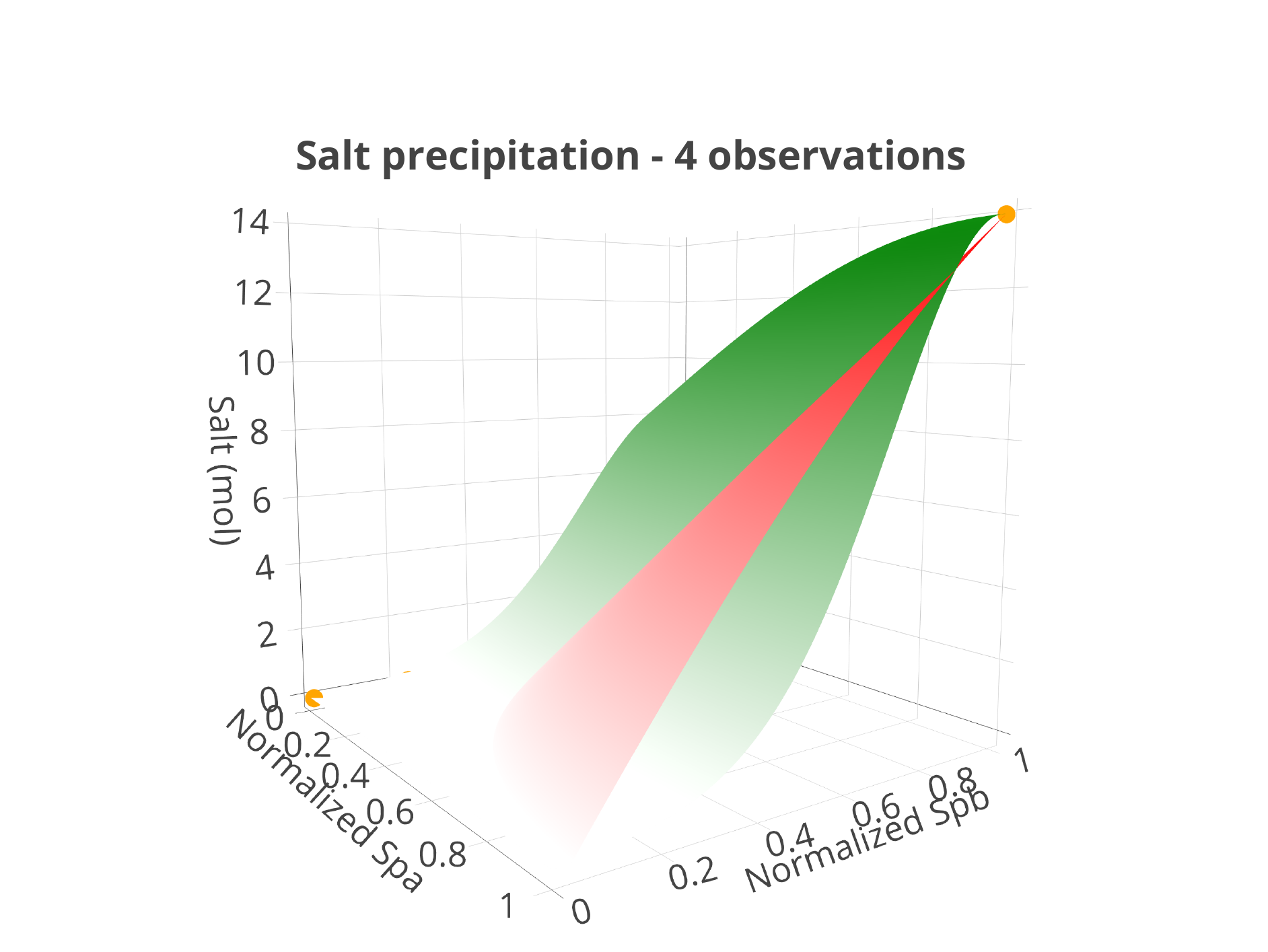}
  \includegraphics[clip, scale=0.3, trim=4.5cm 0.2cm 5cm 3cm]{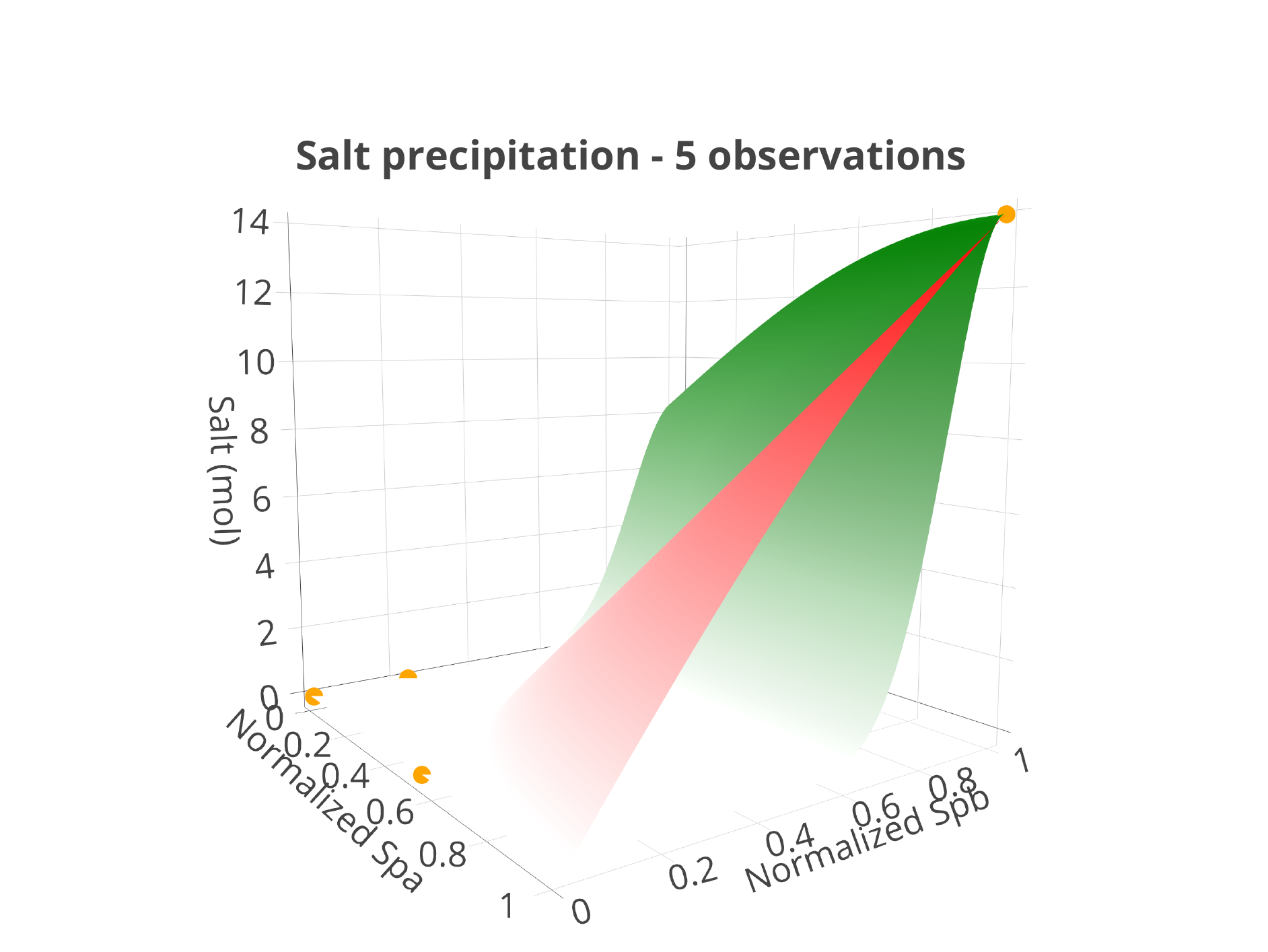} \hspace{1em}
  \includegraphics[clip, scale=0.3, trim=4cm 0.2cm 5cm 3cm]{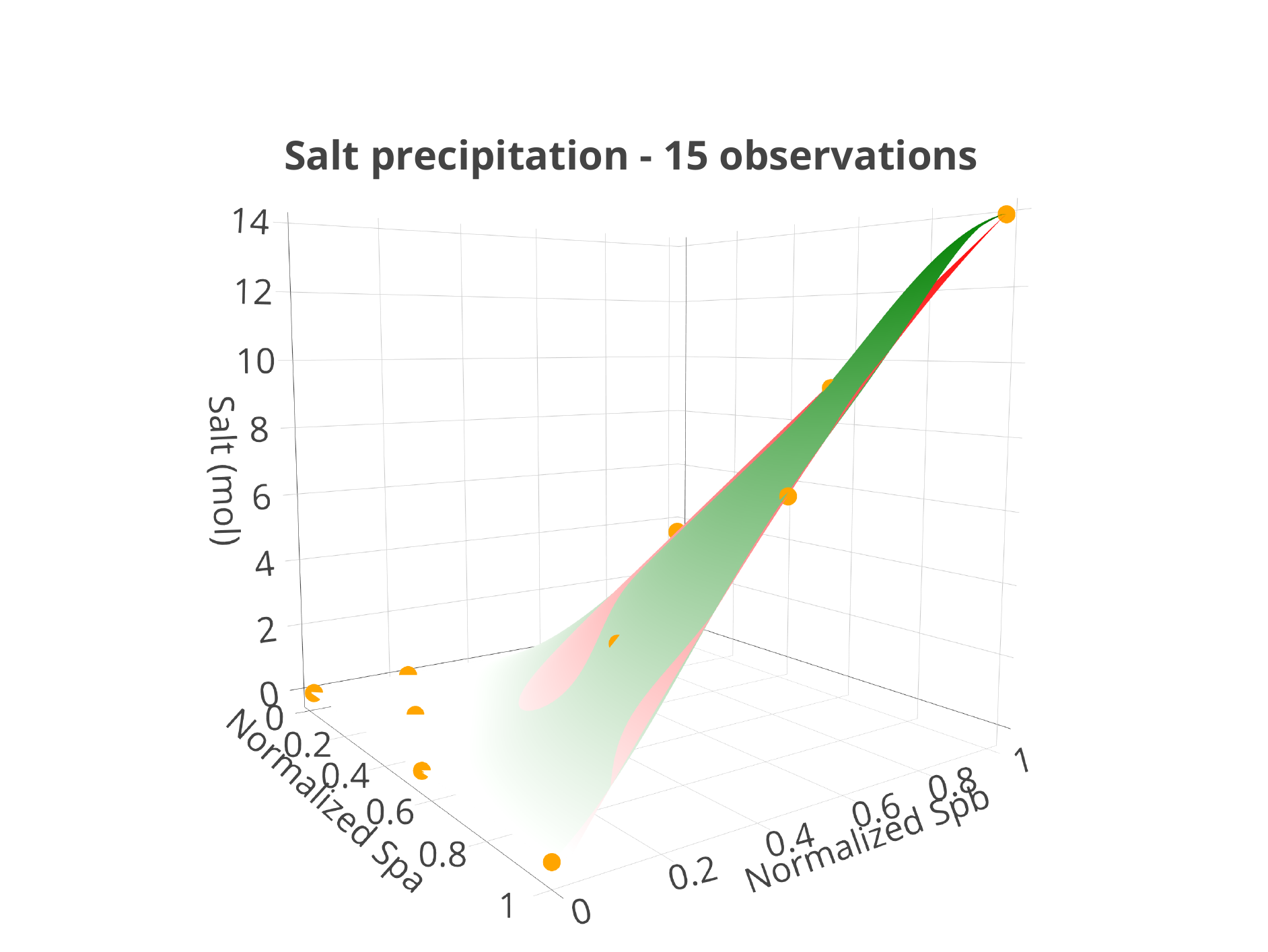}
  \includegraphics[clip, scale=0.3, trim=4.5cm 0.2cm 5cm 3cm]{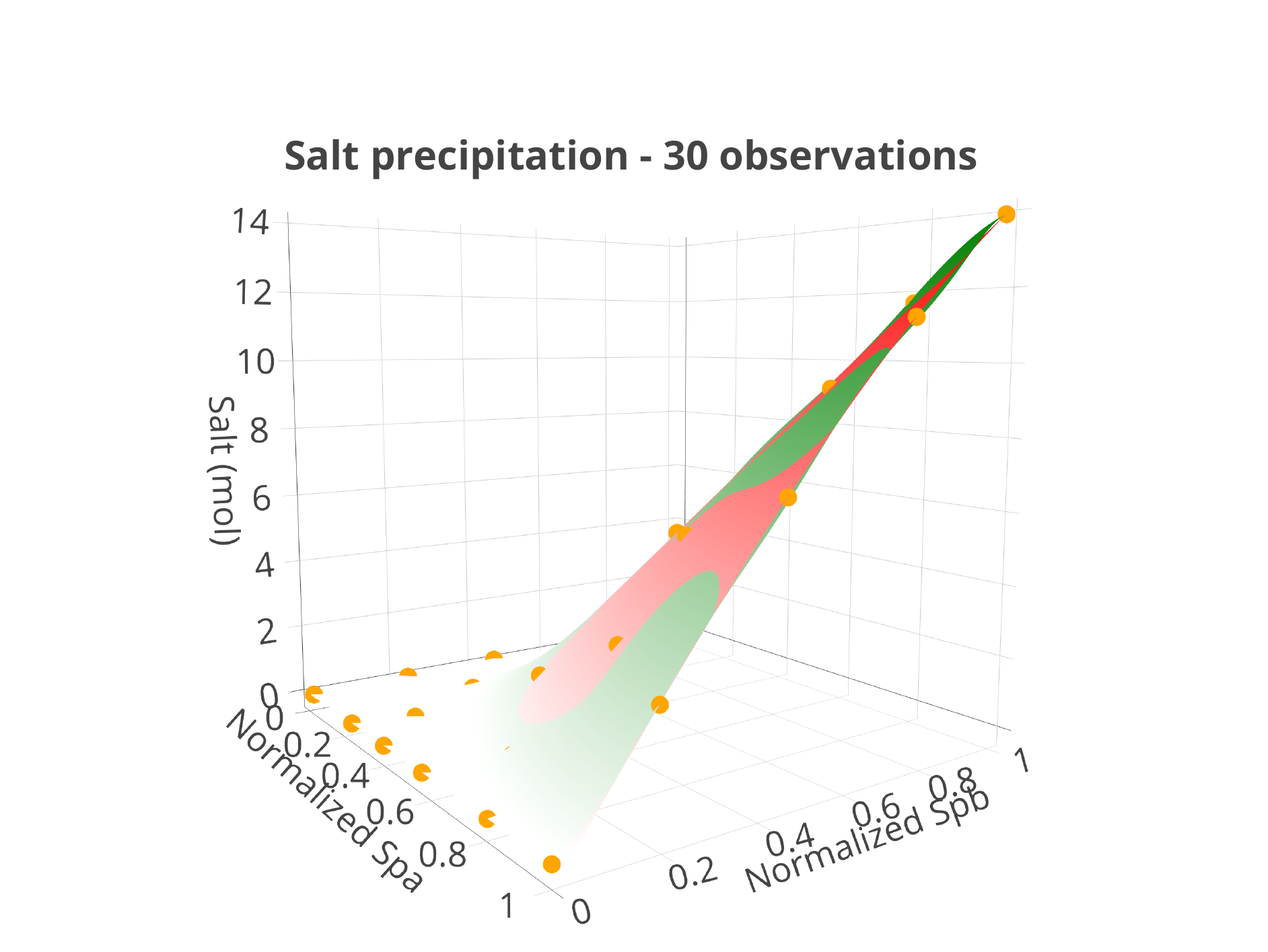} \hspace{1em}
   \includegraphics[clip, scale=0.3, trim=4cm 0.2cm 5cm 3cm]{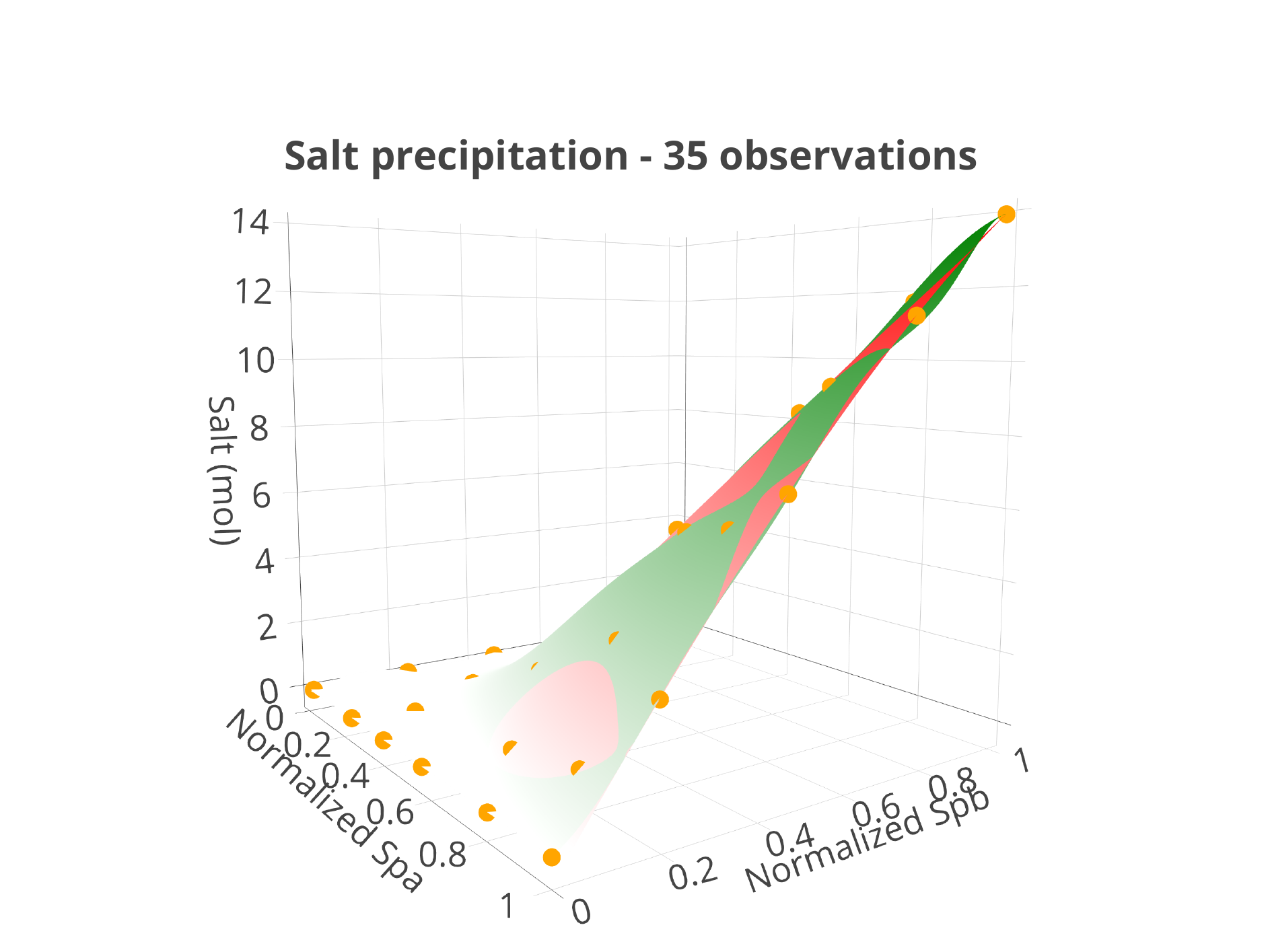}
   \includegraphics[clip, scale=0.4, trim=5cm 7.5cm 3cm 5cm]{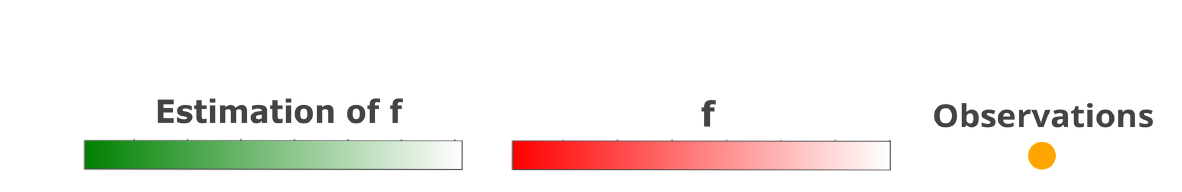} 
  \caption{Illustration of our active learning approach for estimating the function displayed in the right part of Figure \ref{fig:courbes} by starting
    from $t_1=3$ observations randomly chosen in $\textrm{A}\subset [0,1]^2$ for the squared exponential covariance function.
    \label{fig:illustr2D}}
\end{center}
\end{figure}

The different steps of our approach summarized in Algorithm \ref{algo:gps} are illustrated in Figure \ref{fig:illustr2D}.
Here, we used the SE covariance function defined in (\ref{eq:cov_gauss}).

The approach starts with $t_1=3$ points randomly chosen in $\mathrm{A}\subset [0,1]^2$ obtained thanks to a regular grid of $200\times 200$ points.
Then, new points (orange bullets) are added one by one to the set of points at which an evaluation of $f$ is required. These points correspond
at each iteration 
to the position in $\mathrm{A}\subset [0,1]^2$ where the uncertainty $\sigma_{t}^2$ associated to the estimation of $f$ is maximized.
We can see from this figure which displays the true function $f$, the estimation of $f$ and the points at which $f$ has been evaluated that
35 evaluation points are enough to obtain a very accurate estimation of $f$.


\begin{figure}
\begin{center}
  \includegraphics[scale=0.35,trim={0 0 0 1.5cm},clip]{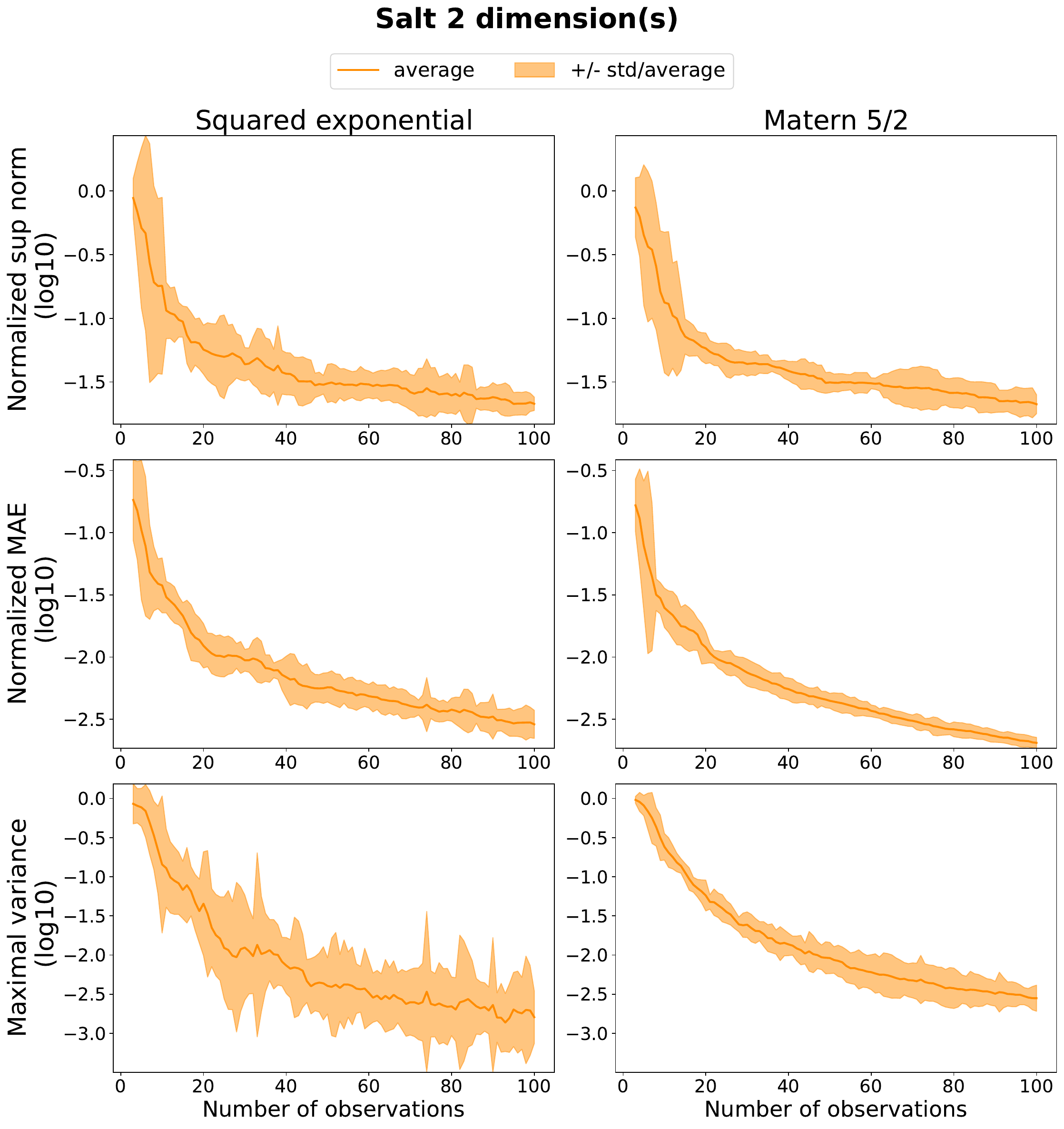}
  \caption{Average and standard deviation of different statistical measures for the squared exponential covariance function defined in
      (\ref{eq:cov_gauss}) (left) and  for the Matern covariance function  defined in (\ref{eq:cov_maternp}) (right) in the case $d=2$. \label{fig:crit2D}}
\end{center}
\end{figure}

In the $d=2$ case, the average and the standard deviation of the statistical measures defined in (\ref{eq:MAE_norm})--(\ref{eq:V})
obtained from 10 replications of the initial set of points are displayed in
Figure \ref{fig:crit2D} for the squared exponential and the Mat\'ern covariance function defined in (\ref{eq:cov_gauss}) and
(\ref{eq:cov_maternp})) for $3\leq t\leq 100$.
We can see that for both choices of covariance function the performance of our approach are
similar: it can reach a normalized  sup norm  (resp. normalized MAE) of $10^{-1.5}$ (resp. $10^{-2.5}$)
by using only $100$ evaluations of the function
to estimate instead of the 40000 points of the grid $\textrm{A}$.
We also observe a smoother behavior of the maximal variance with the Mat\'ern covariance function
even though the final values are close.

\begin{figure}
\begin{center}
\includegraphics[scale=0.4]{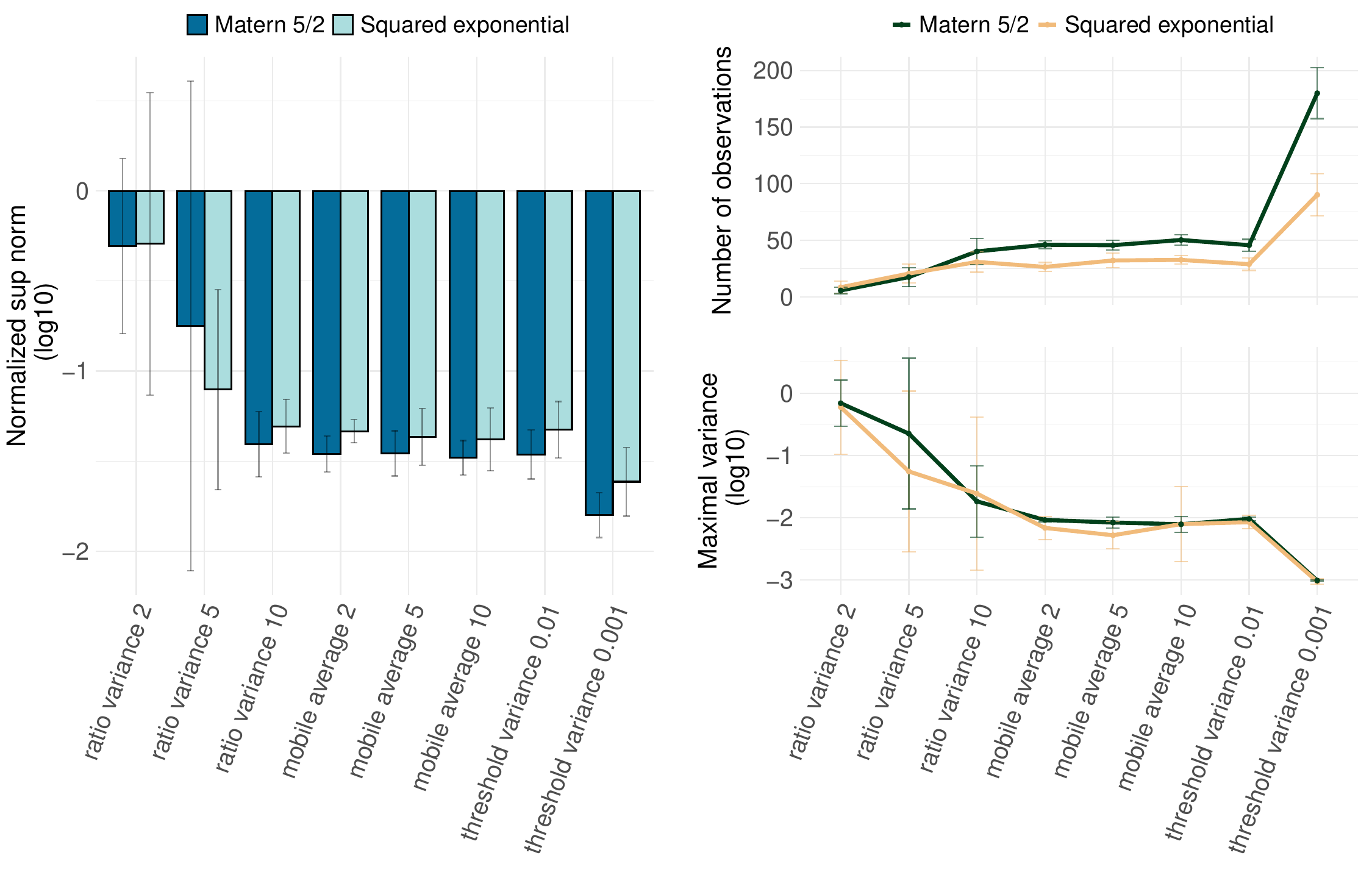}
\caption{Left: Statistical assessment of the error estimation of $f$ displayed in the right part of Figure \ref{fig:courbes} for the stopping criteria
  defined in (\ref{eq:Rk_seuil}), (\ref{eq:Ml_seuil}) and (\ref{eq:V_seuil}) for the squared exponential and the Mat\'ern covariance functions.
  Top right: Number of evaluations required for the different considered stopping criteria. Bottom right: Values of $V(t^\star)$
  where $V$ is defined in (\ref{eq:V}) and $t^\star$ is the stopping iteration which changes from one stopping criterion to another. \label{fig:crit_sum_2D}}
\end{center}
\end{figure}

We can see from the left part of Figure \ref{fig:crit_sum_2D}
that most of the stopping criteria provide an accurate estimation of the function except
``ratio variance 2'' ($R_2$). As for the $d=1$ case, the stopping criteria $R_{10}$ and $M_{10}$ provide very satisfactory results.
Moreover,  we can observe from the right part of this figure that thanks to our active learning approach, 30-50 evaluations of the function to estimate are required
instead of the 40000 points of the initial grid to provide a very accurate
estimation of the function $f$.  Once again, with our approach, we can thus expect a significant reduction of the
computational burden especially in situations where
the computational load associated to the evaluation of $f$ is high.

In this case, the choice of the covariance function might result from a trade-off between accuracy and number of evaluation points.
However, the accuracy and the number of evaluation points do not change drastically suggesting that the choice of the covariance function is still not significant.


%% file: real_revision.tex
The chemical problem solved in this section derives from \cite{kolditz12}. The chemical setup is based on the thermodynamic data for aqueous species and minerals available in the Phreeqc.dat database distributed with PHREEQC \cite{Parkhurst2013}. The compositional system actually solved consists of 14 species in solution, 2 mineral components, 8 geochemical reactions and 2 mineral dissolution-precipitation reactions:

\[\mathrm{H_{2}O  \rightleftharpoons   H^{+} + OH^{-}, logK_1=-13.987}\]
\[\mathrm{HCO_{3}^{-} \rightleftharpoons CO_{3}^{2-} + H^{+},  logK_2=-10.329}\]
\[\mathrm{CO_{2} + H_{2}O \rightleftharpoons CO_{3}^{2-} + 2 H^{+} , logK_3= -16.681 }\]
\[\mathrm{CaHCO_{3}^{+} \rightleftharpoons Ca^{2+} + CO_{3}^{2-} + H^{+}, logK_4=-11.435}\]
\[\mathrm{MgHCO_{3}^{+} \rightleftharpoons Mg^{2+} + H^{+} + CO_{3}^{2-}, logK_5=-11.399 }\]
\[\mathrm{CaCO_{3(aq)} \rightleftharpoons Ca^{2+} + CO_{3}^{2-}, logK_6=-3.224}\]
\[\mathrm{MgCO_{3} \rightleftharpoons Mg^{2+} + CO_{3}^{2-}, logK_7=-2.98 }\]
\[\mathrm{MgOH^{+} + H^{+} \rightleftharpoons Mg^{2+} + H_{2}O, logK_8=11.44}\]
\[\mathrm{Calcite \rightleftharpoons CO_{3}^{2-} + Ca^{2+}, logK_9=-8.48 }\]
\[\mathrm{Dolomite \rightleftharpoons Ca^{2+} + Mg^{2+} + 2 CO_{3}^{2-}, logK_{10}=-17.09}\]

Then, each amount of mineral (calcite or dolomite, respectively) is computed with PHREEQC \cite{Parkhurst2013} as a function of the total elemental concentrations  (C, Ca, Cl, Mg), the pH (as $\mathrm{-log(H^+}$)) and the mineral amount (dolomite or calcite, respectively), which are normalized so that $\mathcal{A}=[0,1]^6$.
Here, our goal is to estimate the functions $f_1$ and $f_2$ defined as follows:
\begin{equation}\label{eq:f_1:f_2}
\textrm{calcite}=f_1(\textrm{C, Ca, Cl, Mg, pH, dolomite}) \textrm{ and } \textrm{dolomite}=f_2(\textrm{C, Ca, Cl, Mg, pH, calcite}),
\end{equation}
by using the minimal number of evaluations of these functions.
For this, we shall use a grid $\mathrm{A}$ built thanks to a Latin Hypercube Sampling (LHS) of $\mathcal{A}$ with $m = 100000$ points.

\begin{figure}
\begin{center}
\includegraphics[clip, scale=0.3, trim=4.5cm 0.2cm 5cm 3cm]{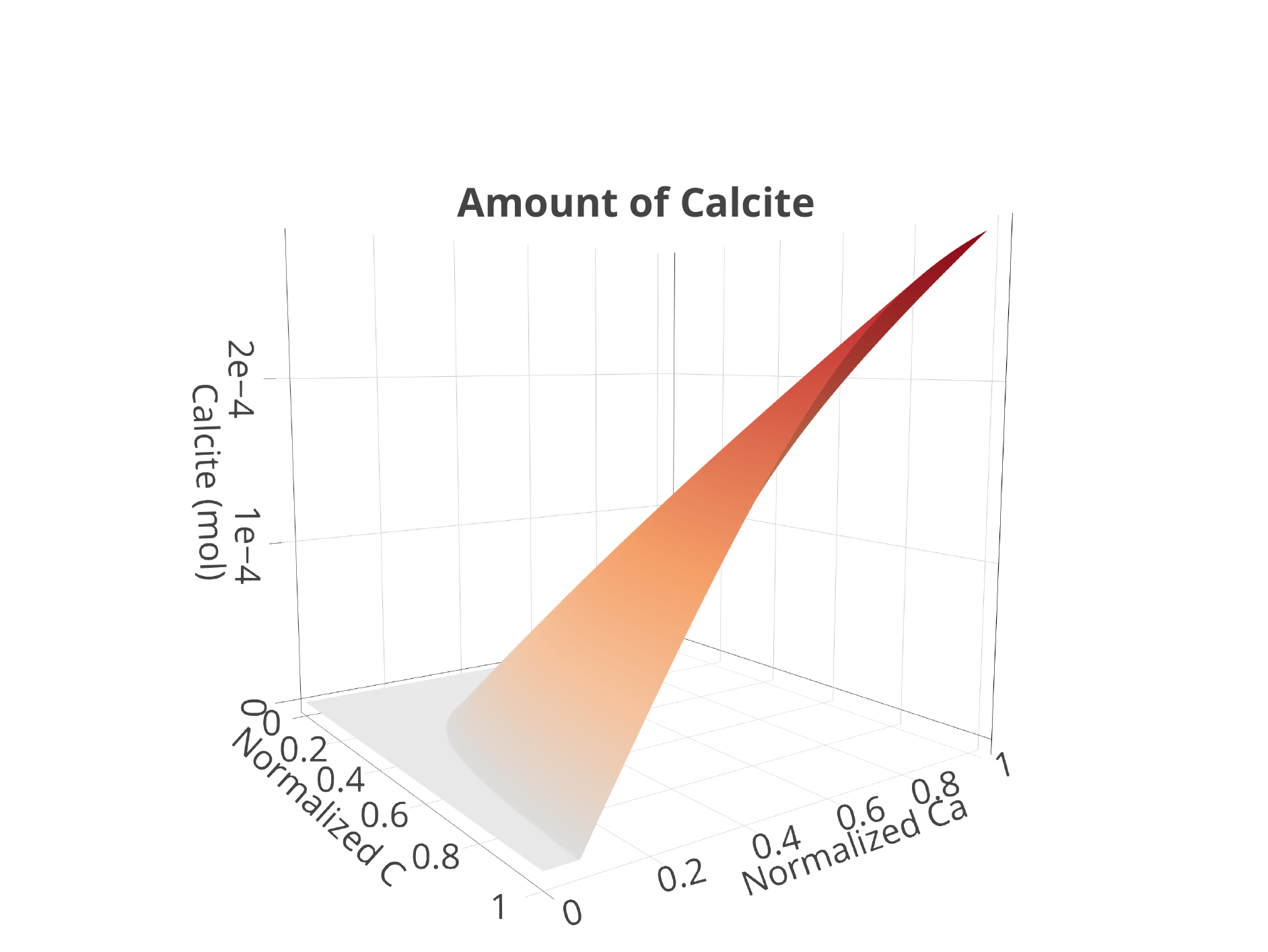}
\includegraphics[clip, scale=0.3, trim=4.5cm 0.2cm 5cm 3cm]{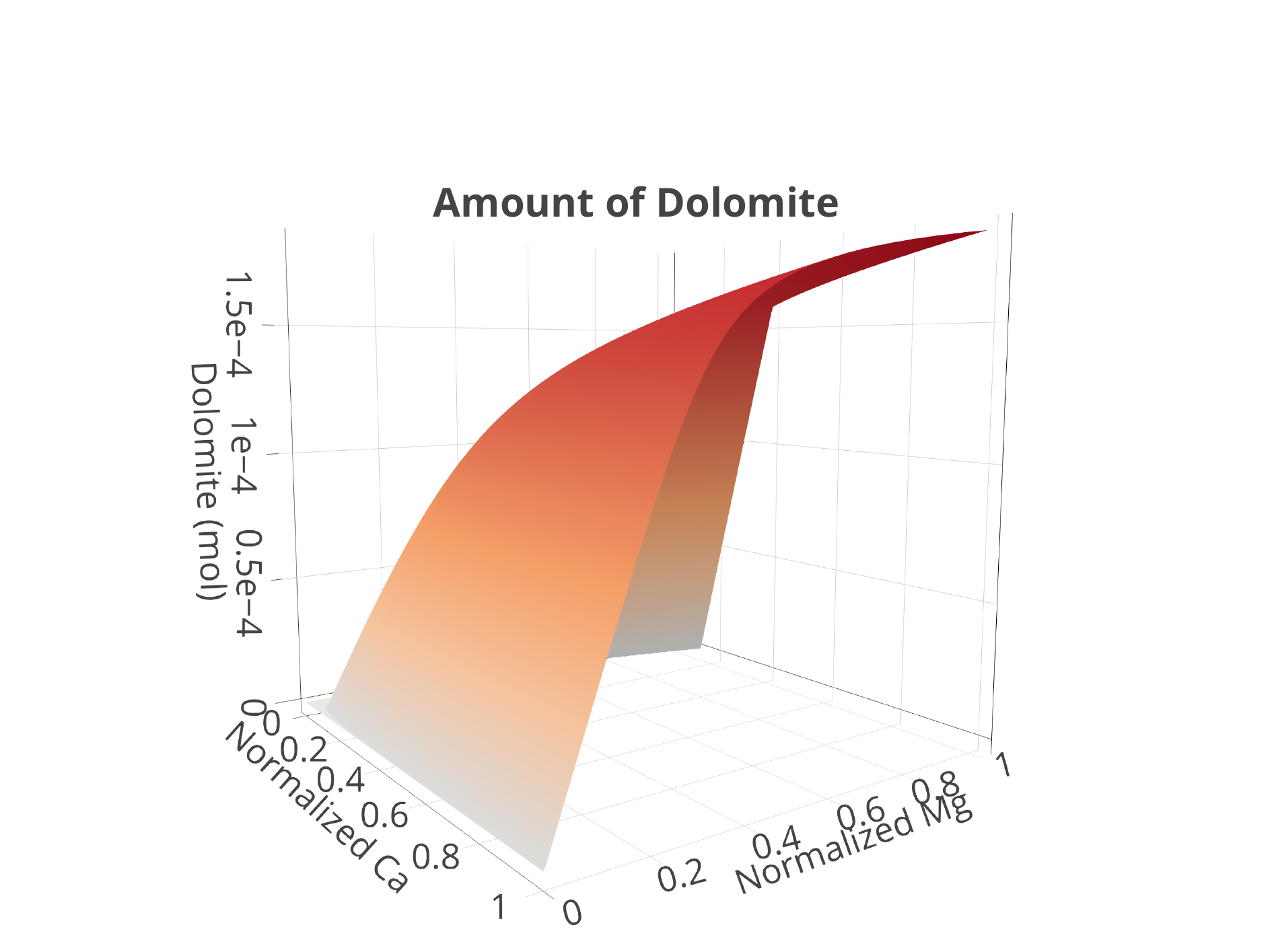}
\caption{Left : Amount of calcite as a function of C and Ca for Cl=$2\times 10^{-3}$ mol/kgw, Mg=$10^{-5}$ mol/kgw, pH=10,
    dolomite=0 mol: $f_1(\textrm{C},\textrm{Ca}, 2\times 10^{-3},10^{-5},10,0)$ where $f_1$ is defined in (\ref{eq:f_1:f_2}).
    Right : Amount of dolomite as a function of Ca and Mg for C=$5\times 10^{-4}$ mol/kgw, Cl=$2\times 10^{-3}$ mol/kgw, pH=10, calcite=0 mol: $f_2(5\times 10^{-4}, \textrm{Ca}, 2\times 10^{-3},\textrm{Mg}, 10, 0)$ where $f_2$ is defined in (\ref{eq:f_1:f_2}).
\label{fig:calcite_dolomite}}
\end{center}
\end{figure}

In the left part of Figure \ref{fig:calcite_dolomite} the amount of calcite is displayed as a function of C and Ca for Cl=$2\times 10^{-3}$ mol/kgw,
  Mg=$10^{-5}$ mol/kgw, pH=10, dolomite=0 mol which corresponds to $f_1(\textrm{C},\textrm{Ca}, 2\times 10^{-3},10^{-5},10,0)$.
  In the right part of Figure \ref{fig:calcite_dolomite} the amount of dolomite is displayed as a function of Ca and  Mg  for C=5$\times 10^{-4}$ mol/kgw,
  Cl=$2\times 10^{-3}$ mol/kgw, pH=10, calcite=0 mol which corresponds to $f_2(5\times 10^{-4}, \textrm{Ca}, 2\times 10^{-3},\textrm{Mg}, 10, 0)$.

  Illustrations of our active learning approach for estimating these functions are shown in Figures \ref{fig:plot_calcite2D} and \ref{fig:plot_dolomite2D}
  of the Appendix.

\subsection{Calcite precipitation}
\begin{figure}
  \includegraphics[scale=0.35,trim={0 0 0 1.5cm},clip]{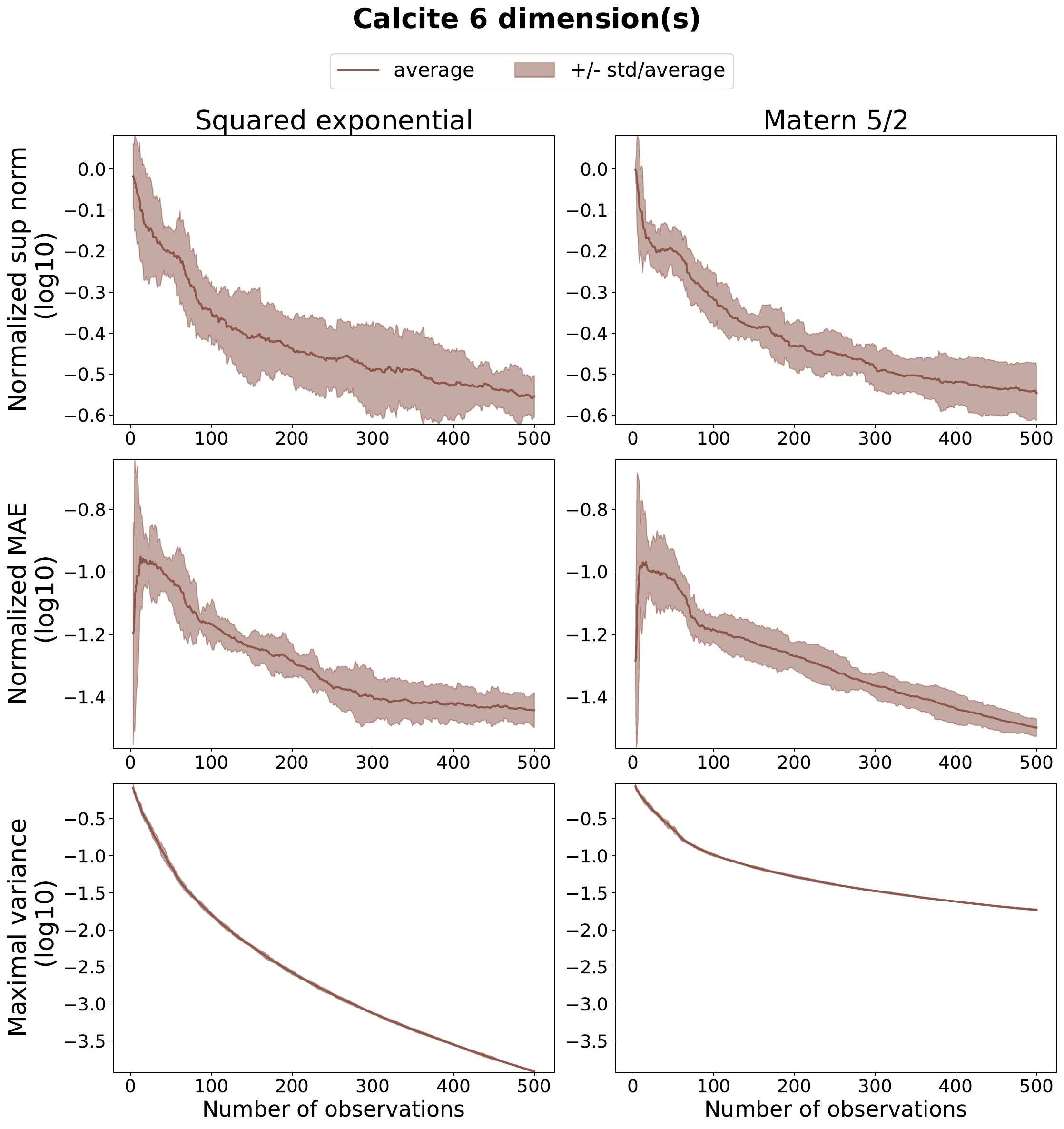}
  \caption{Average and standard deviation of different statistical measures for the squared exponential and the Matérn covariance functions
    defined in (\ref{eq:cov_gauss}) and (\ref{eq:cov_maternp}) for the calcite precipitation problem with $d=6$.\label{fig:calcite_RBF}}
\end{figure}

The average and the standard deviation of the different statistical measures obtained from 10 replications of the initial set of points are shown in Figure \ref{fig:calcite_RBF} for the squared exponential and the Matérn covariance functions defined in (\ref{eq:cov_gauss}) and (\ref{eq:cov_maternp}) for $3\leq t \leq 500$. We can see that for both choices of covariance functions, the maximal variance and the statistical precision measures keep decreasing as the number of evaluations increases. For instance, our method allows us to have a normalized  sup norm (resp. normalized MAE) of $10^{-0.5}$ (resp. $10^{-1.4}$) with only 500 evaluations instead of the 100 000 points of the grid $\textrm{A}$ for both covariance functions. However, the maximal variance is around
$10^{-3.5}$ (resp. $10^{-1.5}$) for the squared exponential (resp. Mat\'ern) covariance function.

\begin{figure}
  \includegraphics[scale=0.4]{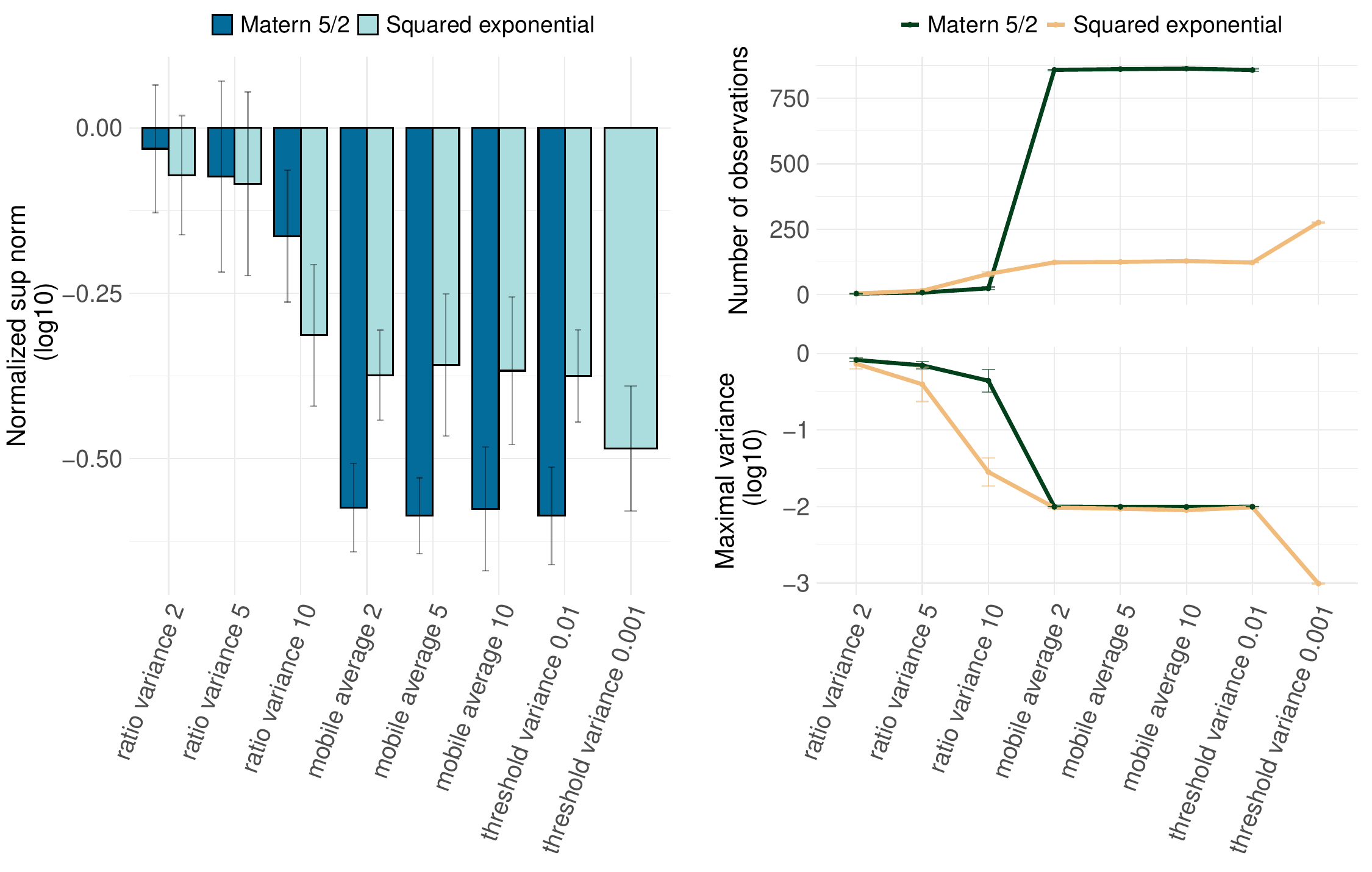}
  \caption{Left: Statistical assessment of the error estimation of $f_1$ defined in (\ref{eq:f_1:f_2}) for the stopping criteria
    defined in (\ref{eq:Rk_seuil}), (\ref{eq:Ml_seuil}) and (\ref{eq:V_seuil}) for the squared exponential and the Matérn covariance function
    defined in (\ref{eq:cov_gauss}) and (\ref{eq:cov_maternp}).
  Top right: Number of evaluations required for the different considered stopping criteria. Bottom right: Values of $V(t^\star)$
  where $V$ is defined in (\ref{eq:V}) and $t^\star$ is the stopping iteration which changes from one stopping criterion to another.\label{fig:calcite_sum_RBF}}
\end{figure}

Moreover, we can see from Figure \ref{fig:calcite_sum_RBF} that when the mobile average $M_\ell$ criteria and the squared exponential covariance function are used
  the final estimation of $f_1$ is obtained with around 100 evaluations of $f_1$  instead of $10^5$.
  To obtain similar statistical performance with the Mat\'ern covariance more than 750 observations are required.
  The difference between the two covariance functions probably comes from the behavior of the maximal variance. It is still strongly decreasing
  after 500 observations
  for the squared exponential covariance function which is not the case for the Mat\'ern covariance function.

\subsection{Dolomite precipitation}
\begin{figure}
  \includegraphics[scale=0.35,trim={0 0 0 1.5cm},clip]{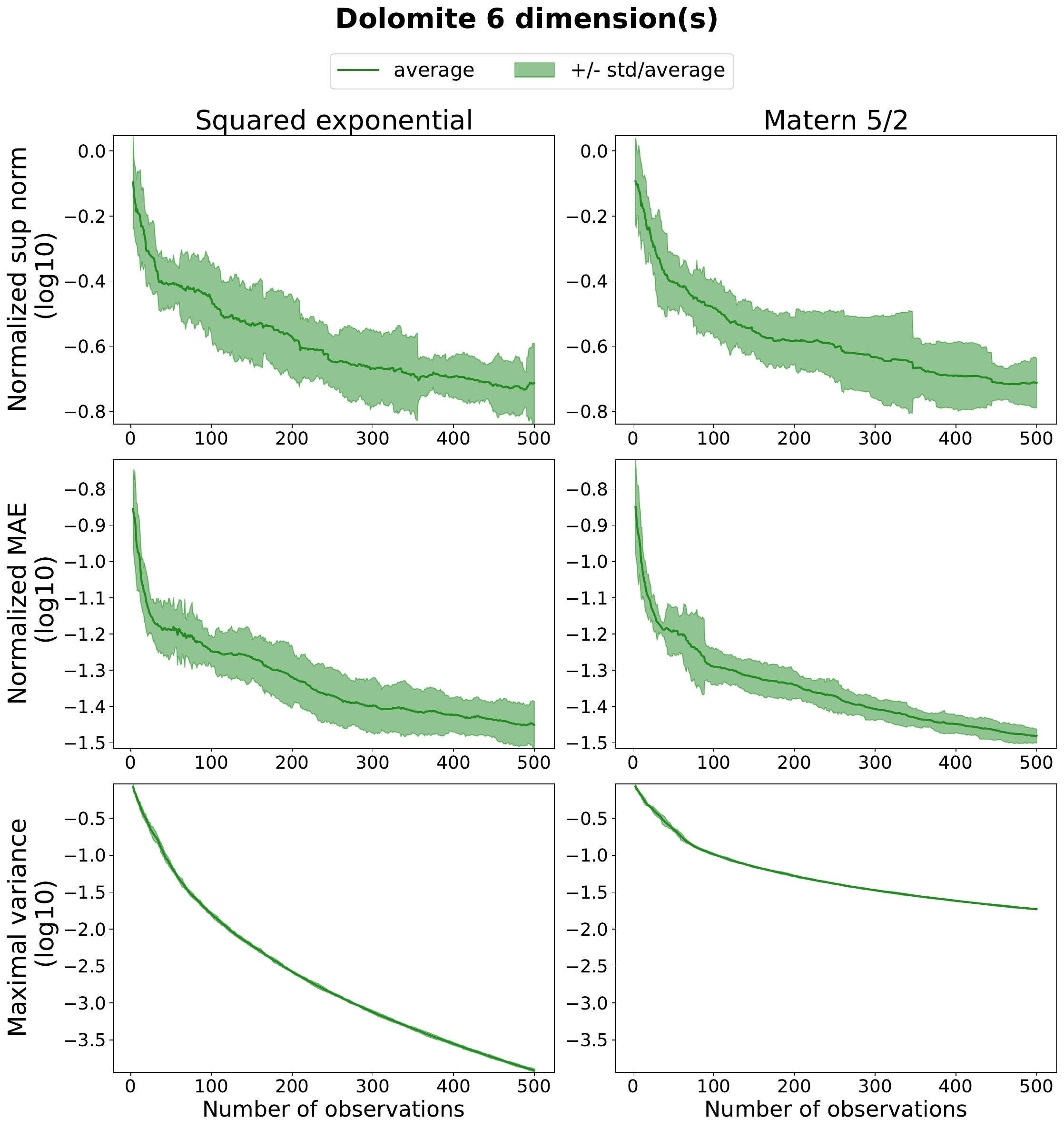}
  \caption{Average and standard deviation of different statistical measures for the squared exponential and the Mat\'ern covariance functions
    defined in (\ref{eq:cov_gauss}) and (\ref{eq:cov_maternp}) for the dolomite precipitation problem with $d=6$.\label{fig:dolomite_RBF}}
\end{figure}

\begin{figure}
  \includegraphics[scale=0.4]{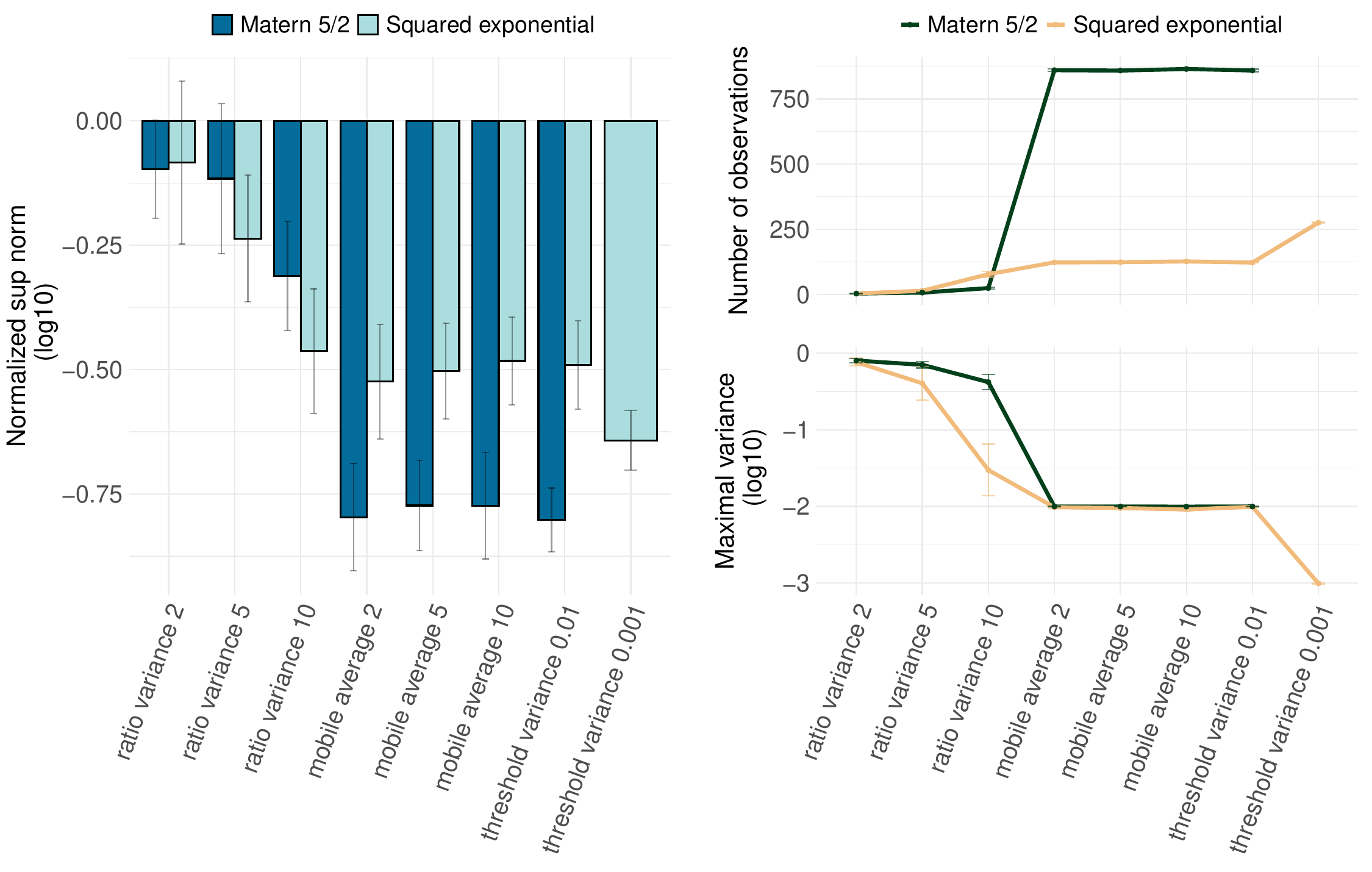}
  \caption{Left: Statistical assessment of the error estimation of $f_2$ defined in (\ref{eq:f_1:f_2}) for the stopping criteria
  defined in (\ref{eq:Rk_seuil}), (\ref{eq:Ml_seuil}) and (\ref{eq:V_seuil}) for the squared exponential and the Mat\'ern covariance functions
    defined in (\ref{eq:cov_gauss}) and (\ref{eq:cov_maternp}).
  Top right: Number of evaluations required for the different considered stopping criteria. Bottom right: Values of $V(t^\star)$
  where $V$ is defined in (\ref{eq:V}) and $t^\star$ is the stopping iteration which changes from one stopping criterion to another.\label{fig:dolomite_sum_RBF}}
\end{figure}

Similarly to the previous case, the average and the standard deviation of the different statistical measures obtained from 10 replications of the initial set of points are shown in Figure \ref{fig:dolomite_RBF} for the squared exponential and the Matérn covariance functions defined in (\ref{eq:cov_gauss}) and (\ref{eq:cov_maternp}) for $3\leq t \leq 500$. We obtained similar conclusions as for the calcite precipitation case, see Figure \ref{fig:dolomite_sum_RBF}.



%% file: conclusion_revision.tex
We have shown that our method has two main features which make it very attractive. Firstly, it is very efficient from a practical point of view 
thanks to the Gaussian Process modeling which enables us to sequentially build the surrogate
model with a low number of points and almost no parameters to tune. Secondly, its very low computational burden makes its use possible
on complex chemical reactions involving singular behaviors like precipitation and dissolution of minerals.
Our method could also be applied to more complex geochemical systems like
  surface complexation or ion exchange that can be described with laws of mass action.
Effectively, these two features have further potential applications on much larger sets of reactive species or
with coupled physical processes namely in reactive transport modeling.
This will be the subject of a future work.


%% file: appendix_revision.tex
\begin{figure}
\begin{center}
 \includegraphics[clip, scale=0.3, trim=4.5cm 0.2cm 5cm 3cm]{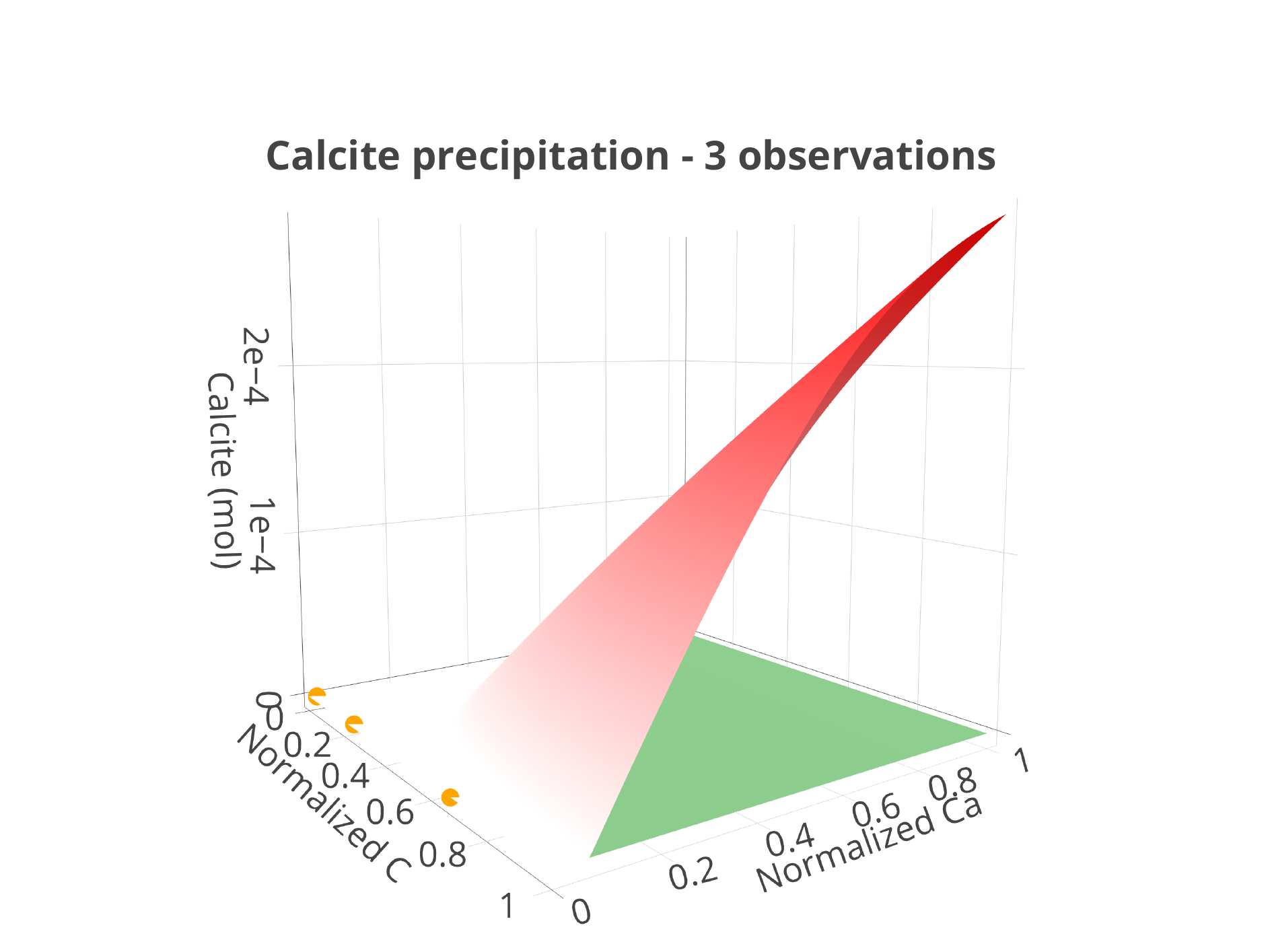} \hspace{1em}
  \includegraphics[clip, scale=0.3, trim=4cm 0.2cm 5cm 3cm]{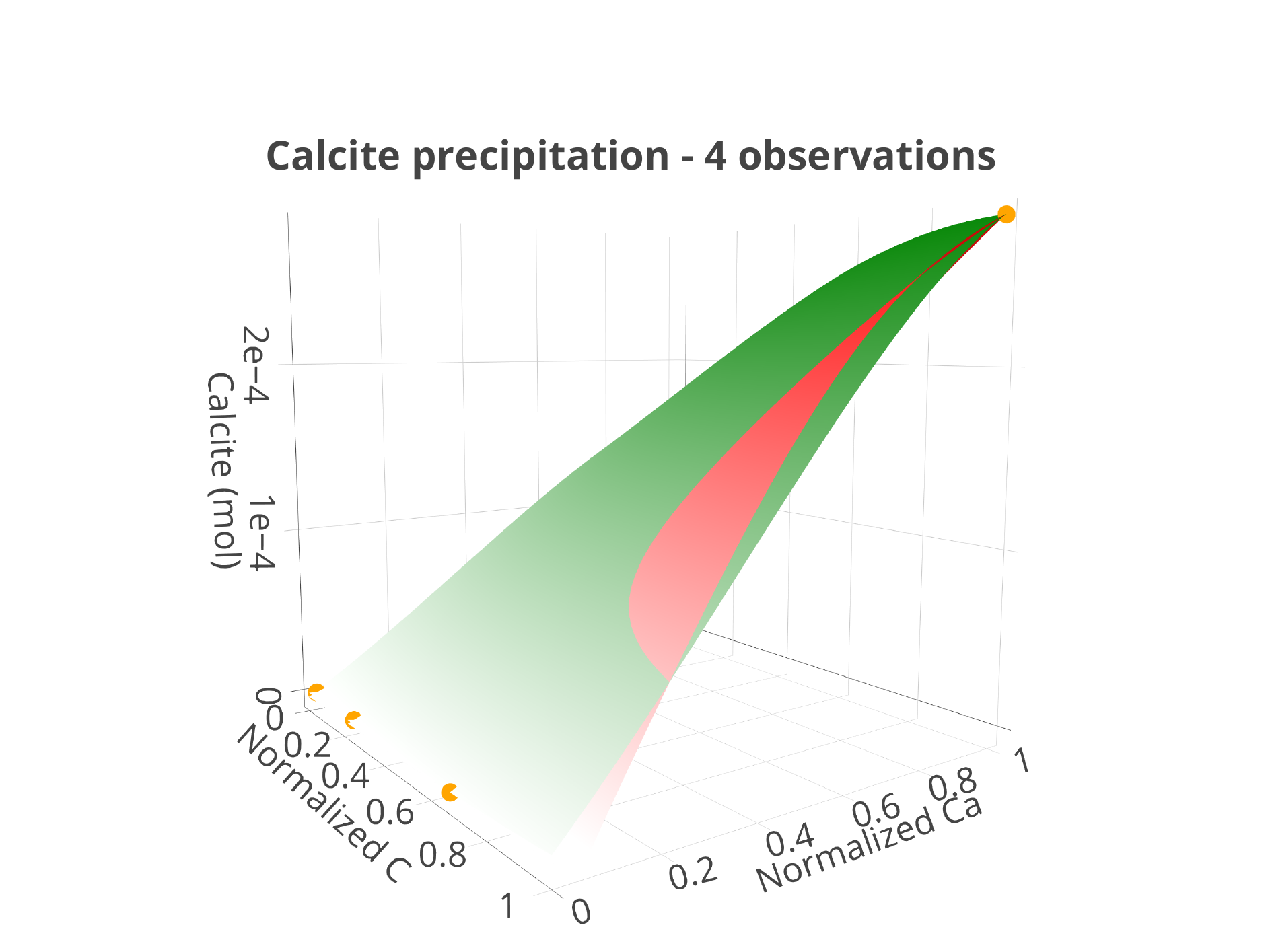}
  \includegraphics[clip, scale=0.3, trim=4.5cm 0.2cm 5cm 3cm]{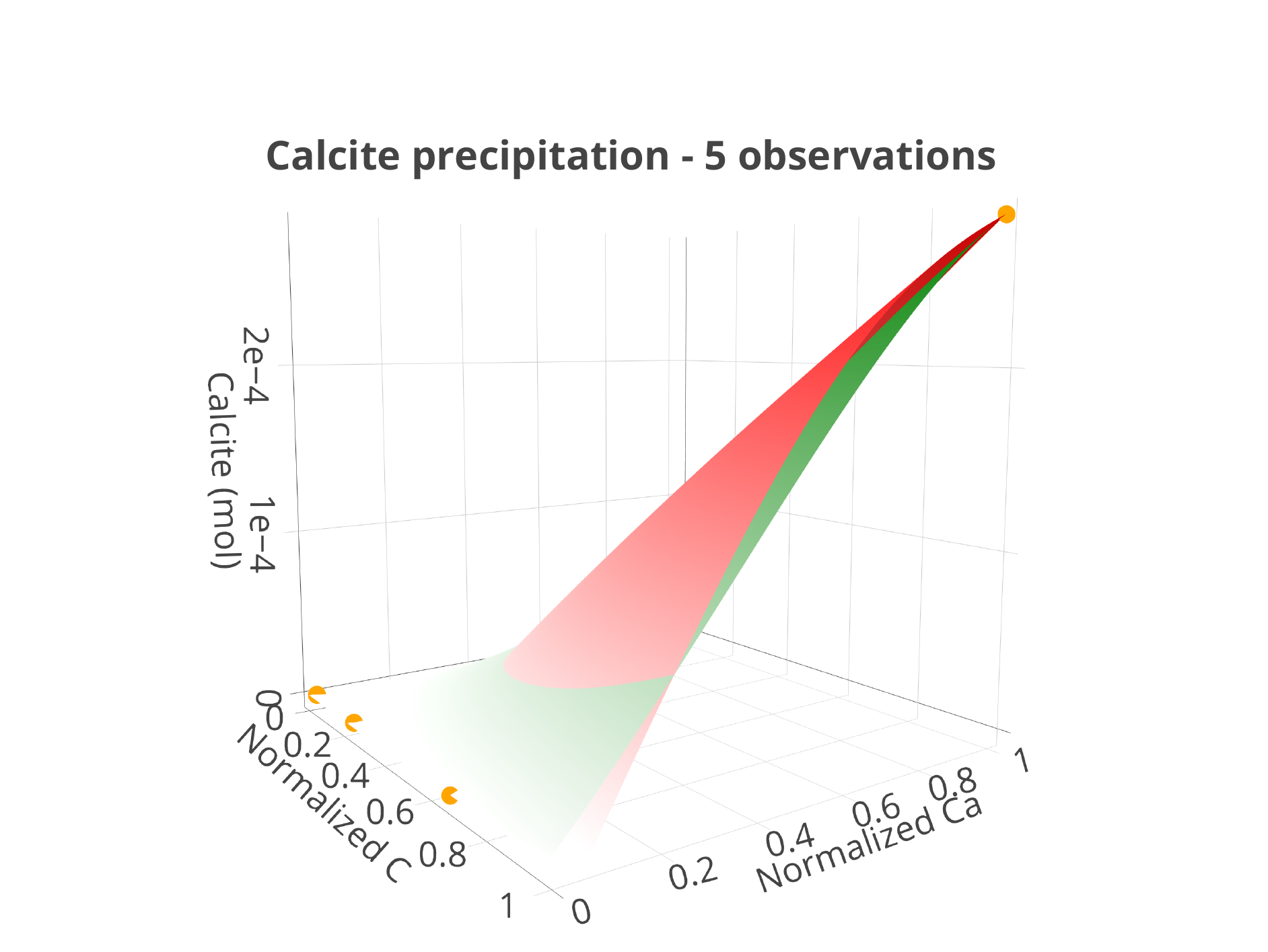} \hspace{1em}
  \includegraphics[clip, scale=0.3, trim=4cm 0.2cm 5cm 3cm]{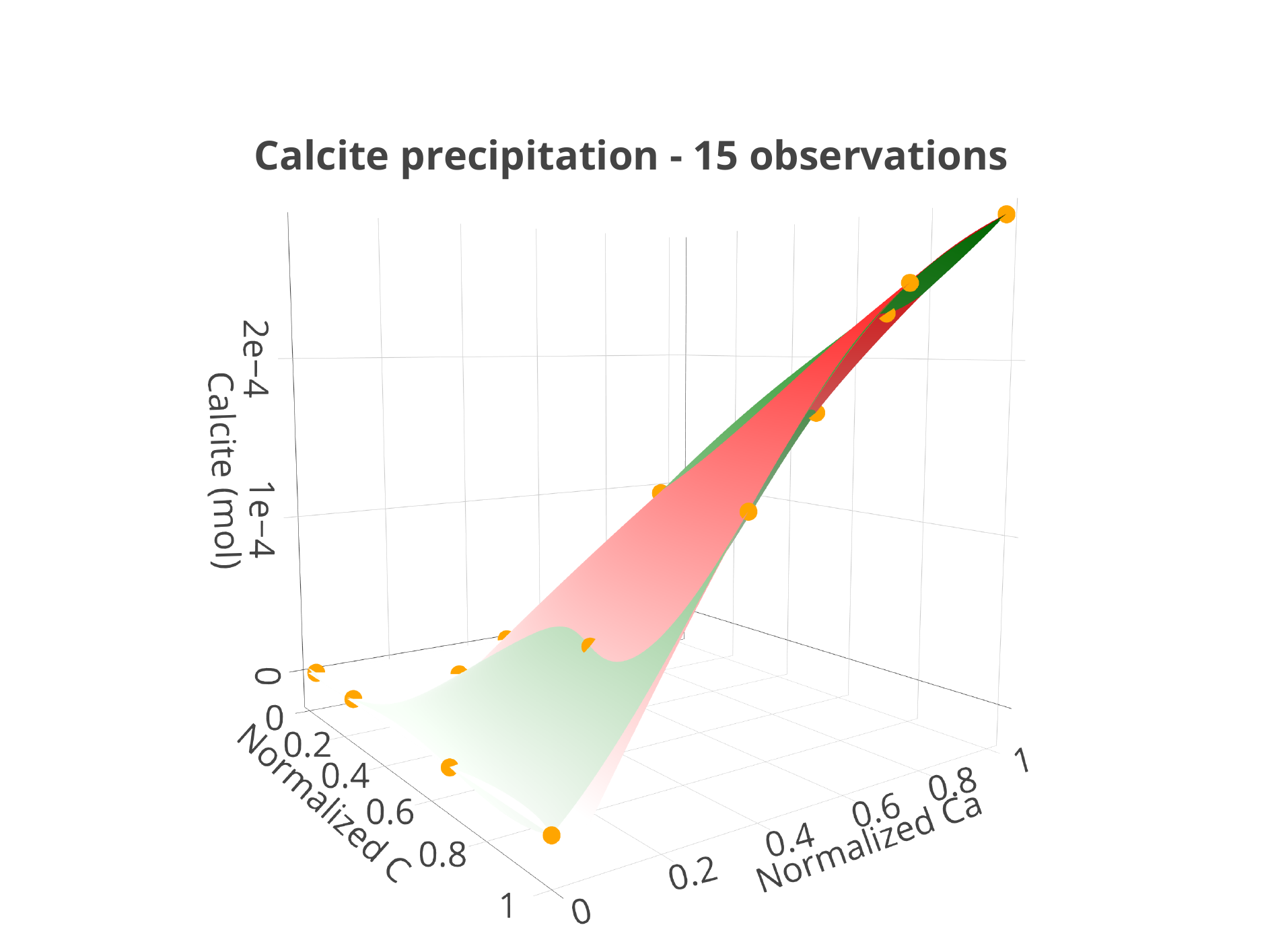}
  \includegraphics[clip, scale=0.3, trim=4.5cm 0.2cm 5cm 3cm]{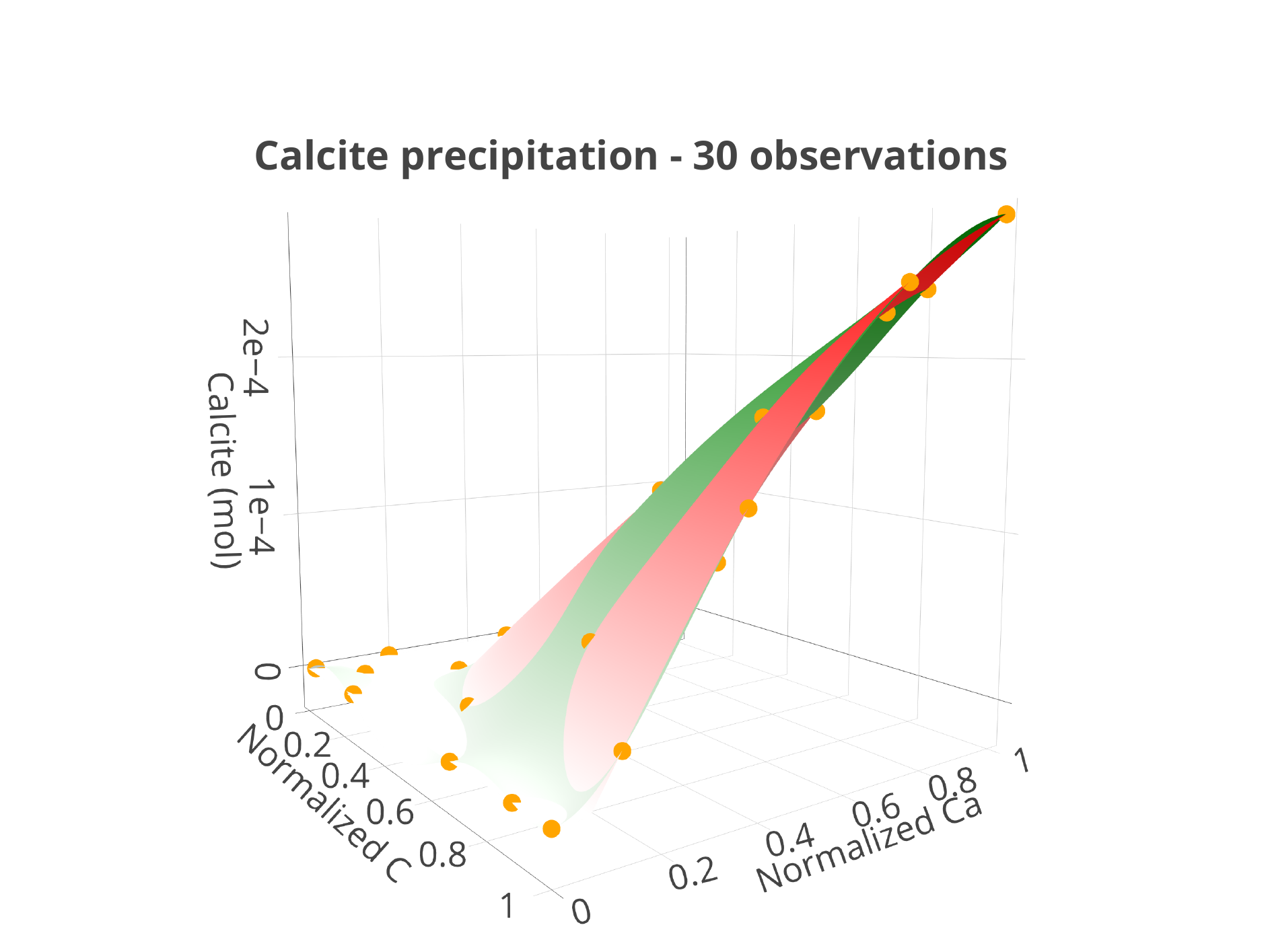} \hspace{1em}
   \includegraphics[clip, scale=0.3, trim=4cm 0.2cm 5cm 3cm]{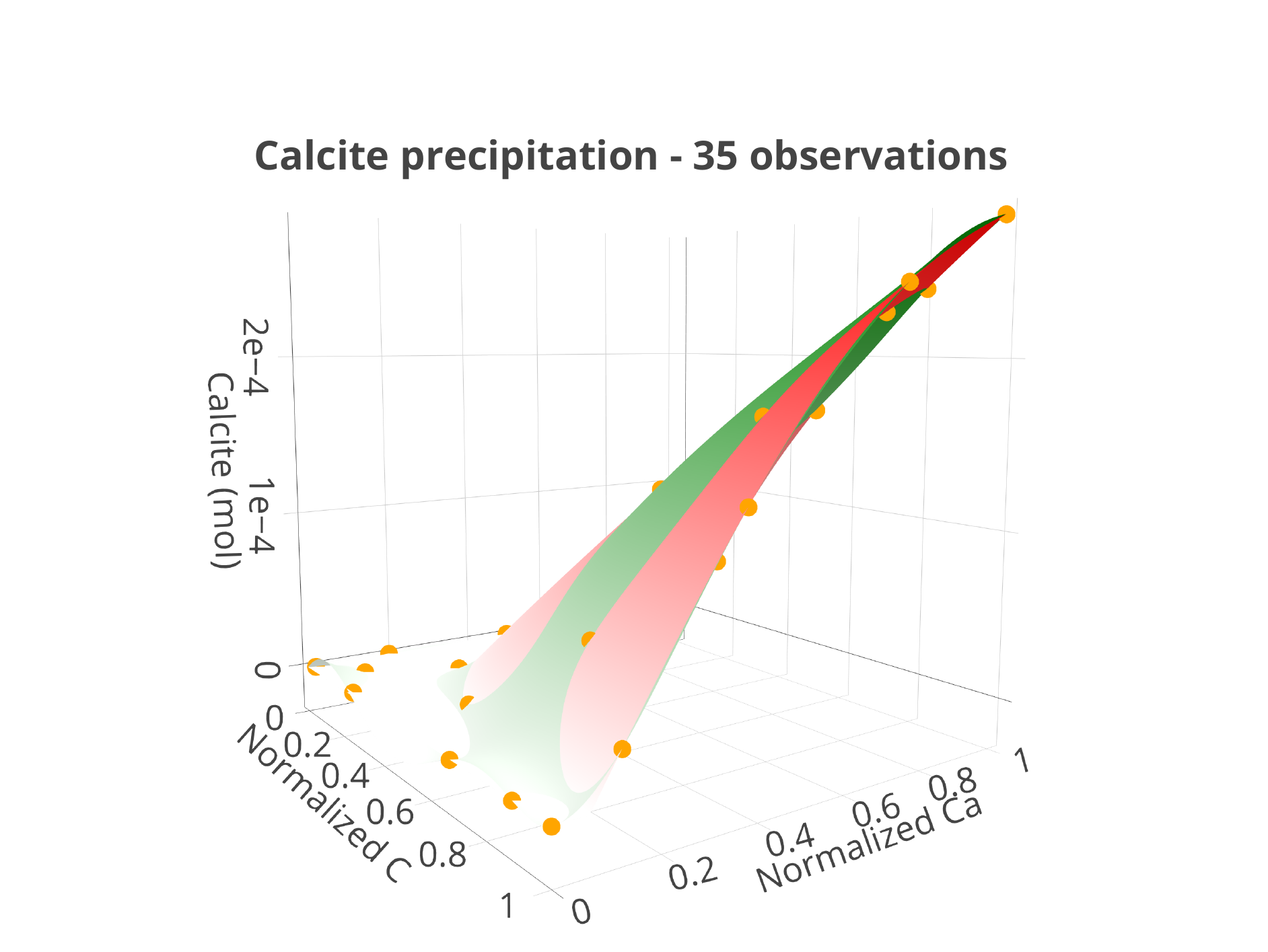}
   \includegraphics[clip, scale=0.4, trim=5cm 7.5cm 3cm 5cm]{Fig/visualisation_courbes/scale_2D.pdf} 
  \caption{Illustration of our active learning approach for estimating the function $f_1(\textrm{C},\textrm{Ca}, 2\times 10^{-3},10^{-5},10,0)$ displayed in the left part of Figure \ref{fig:calcite_dolomite} by starting
  from $t_1=3$ observations randomly chosen in $\textrm{A}\subset [0,1]^2$. Here, the squared exponential covariance function was used.\label{fig:plot_calcite2D}}
\end{center}
\end{figure}

\begin{figure}
\begin{center}
 \includegraphics[clip, scale=0.3, trim=4.5cm 0.2cm 5cm 3cm]{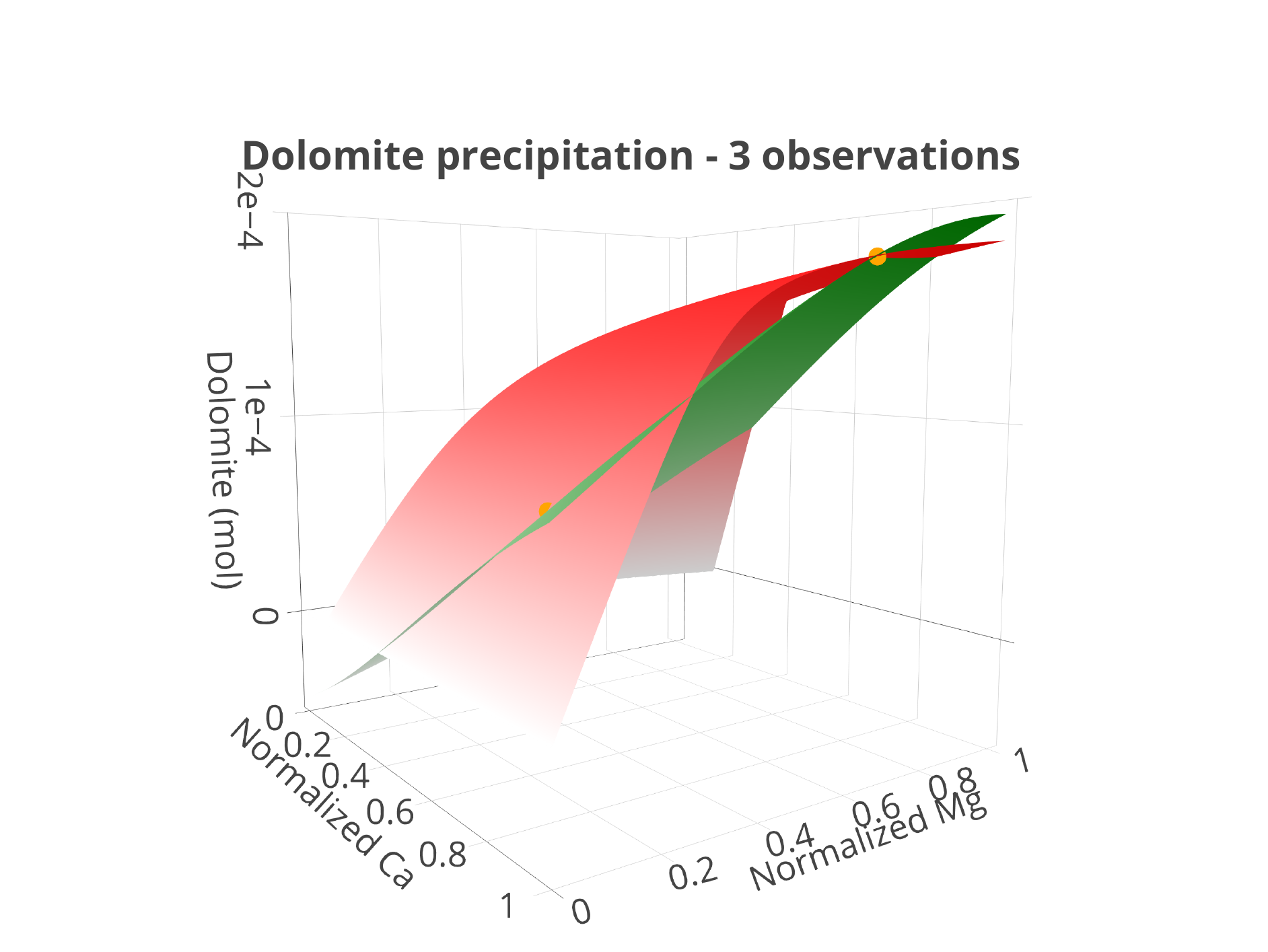} \hspace{1em}
  \includegraphics[clip, scale=0.3, trim=4cm 0.2cm 5cm 3cm]{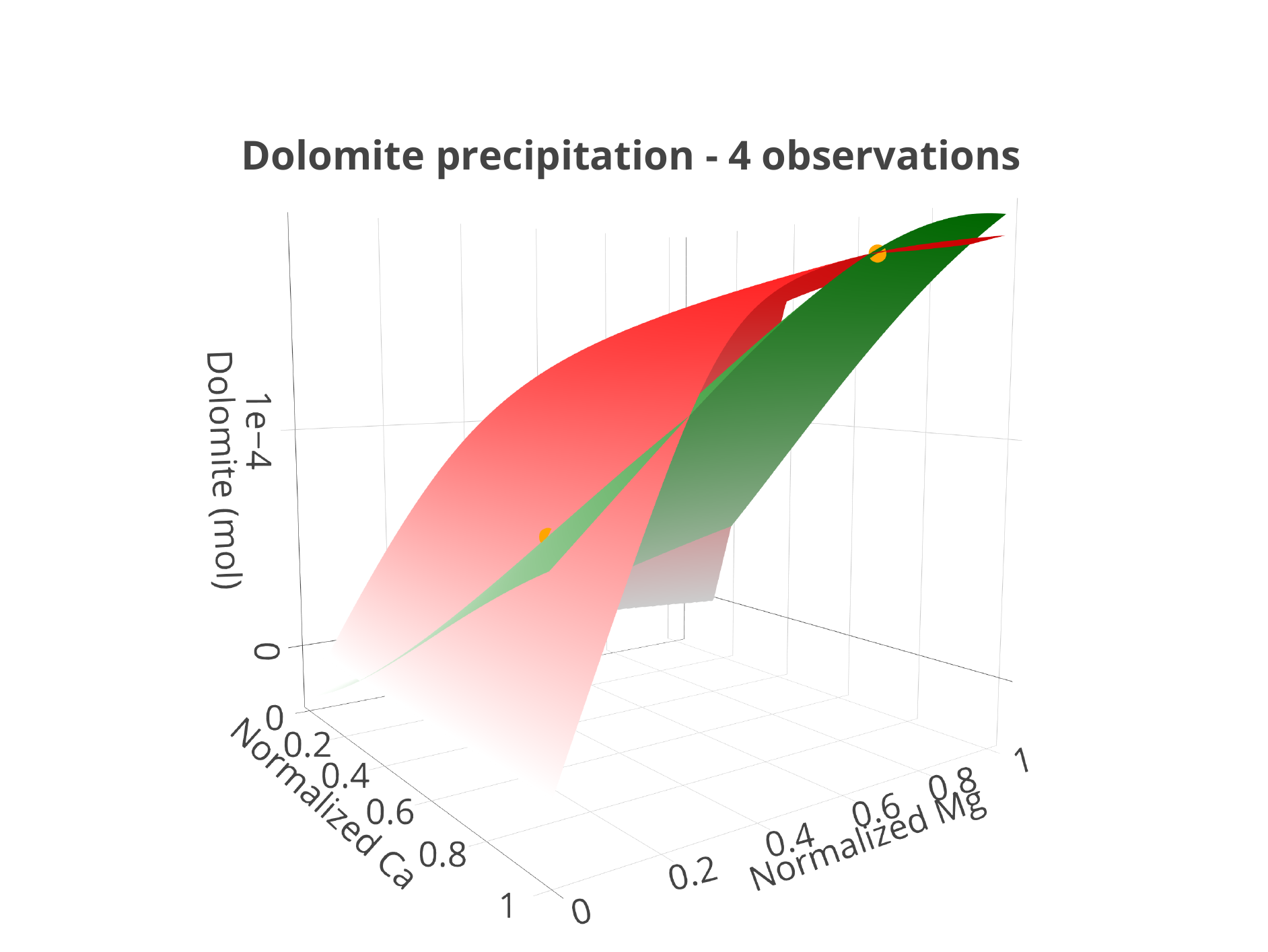}
  \includegraphics[clip, scale=0.3, trim=4.5cm 0.2cm 5cm 3cm]{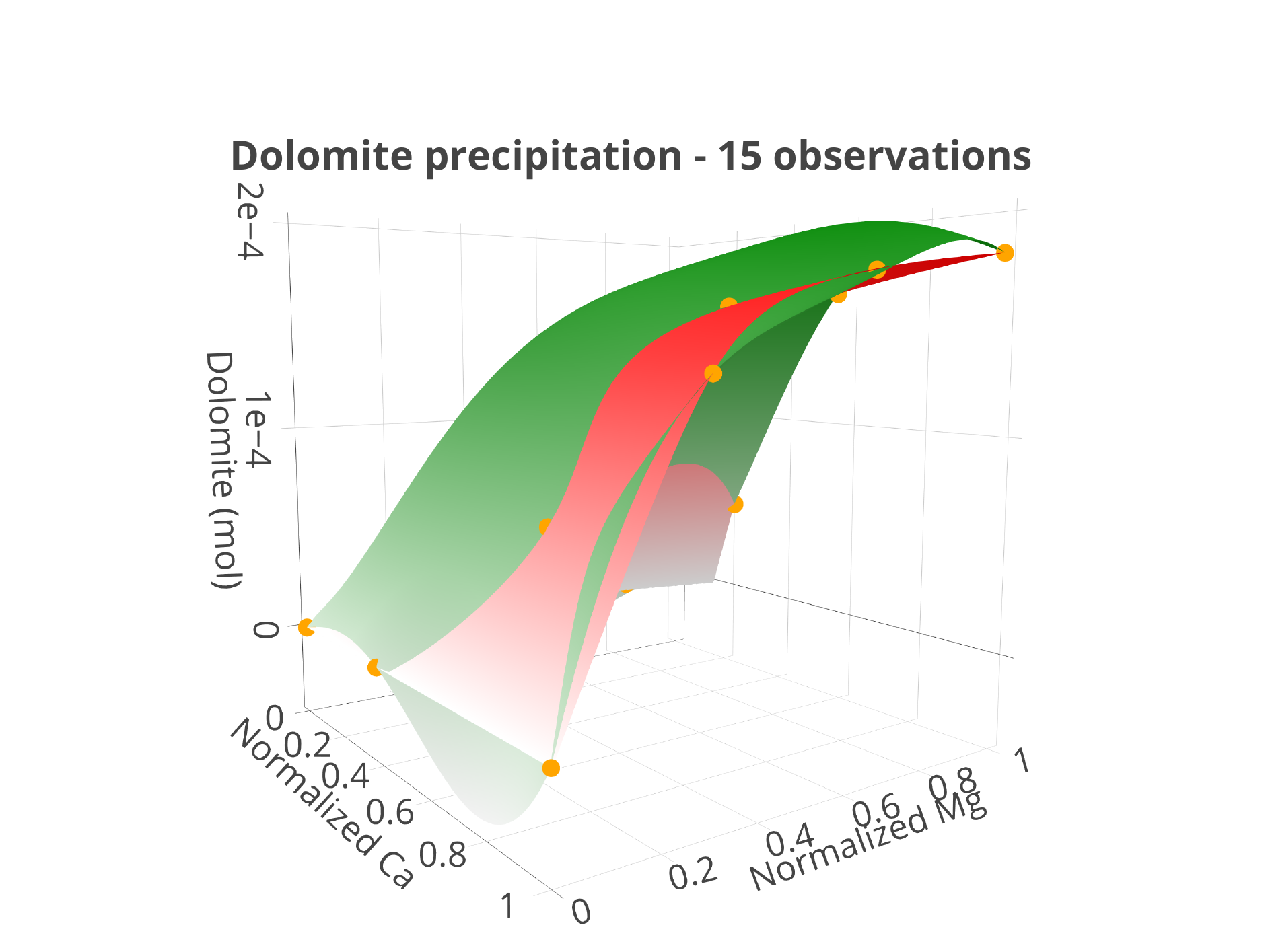} \hspace{1em}
  \includegraphics[clip, scale=0.3, trim=4cm 0.2cm 5cm 3cm]{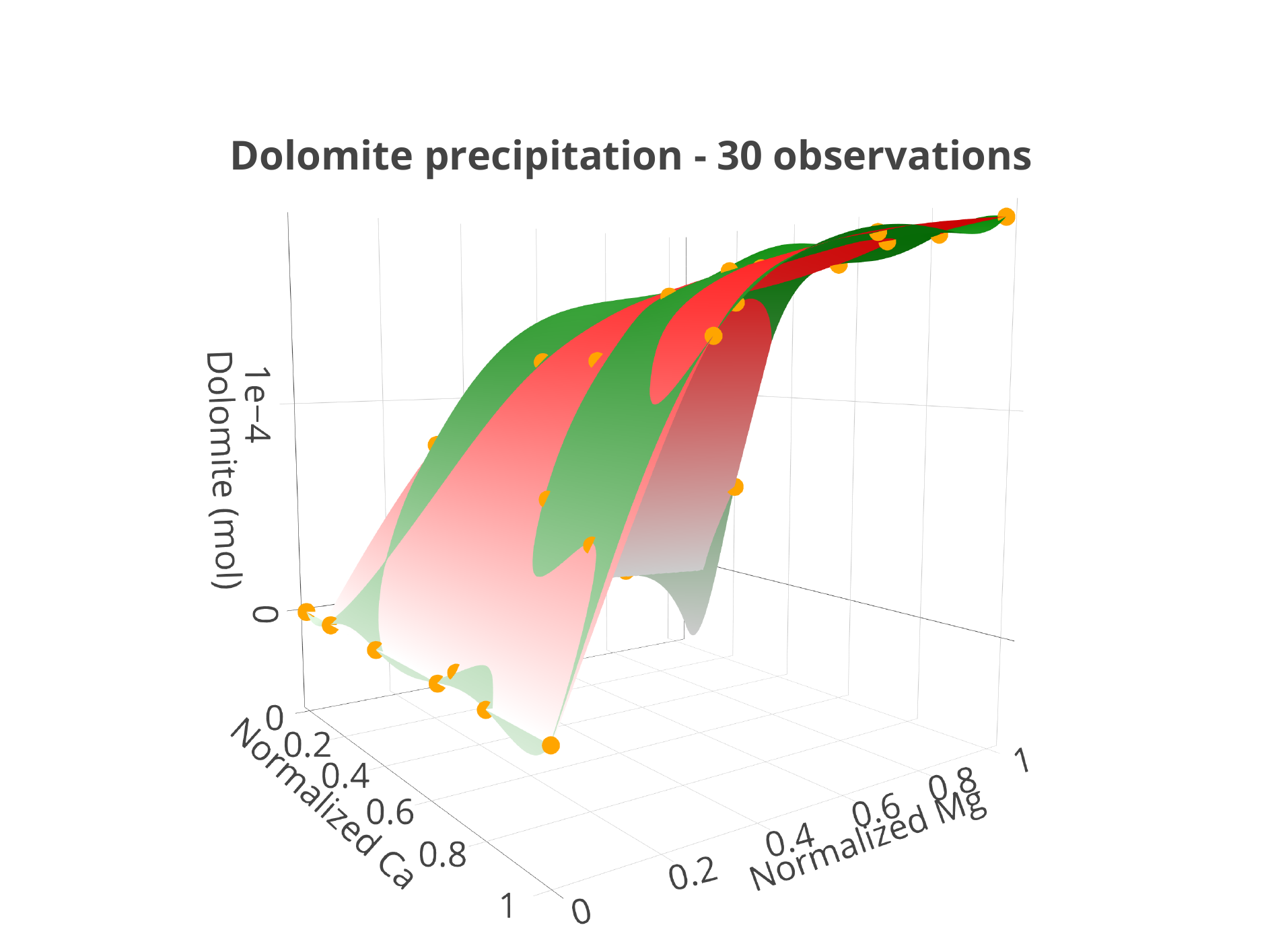}
  \includegraphics[clip, scale=0.3, trim=4.5cm 0.2cm 5cm 3cm]{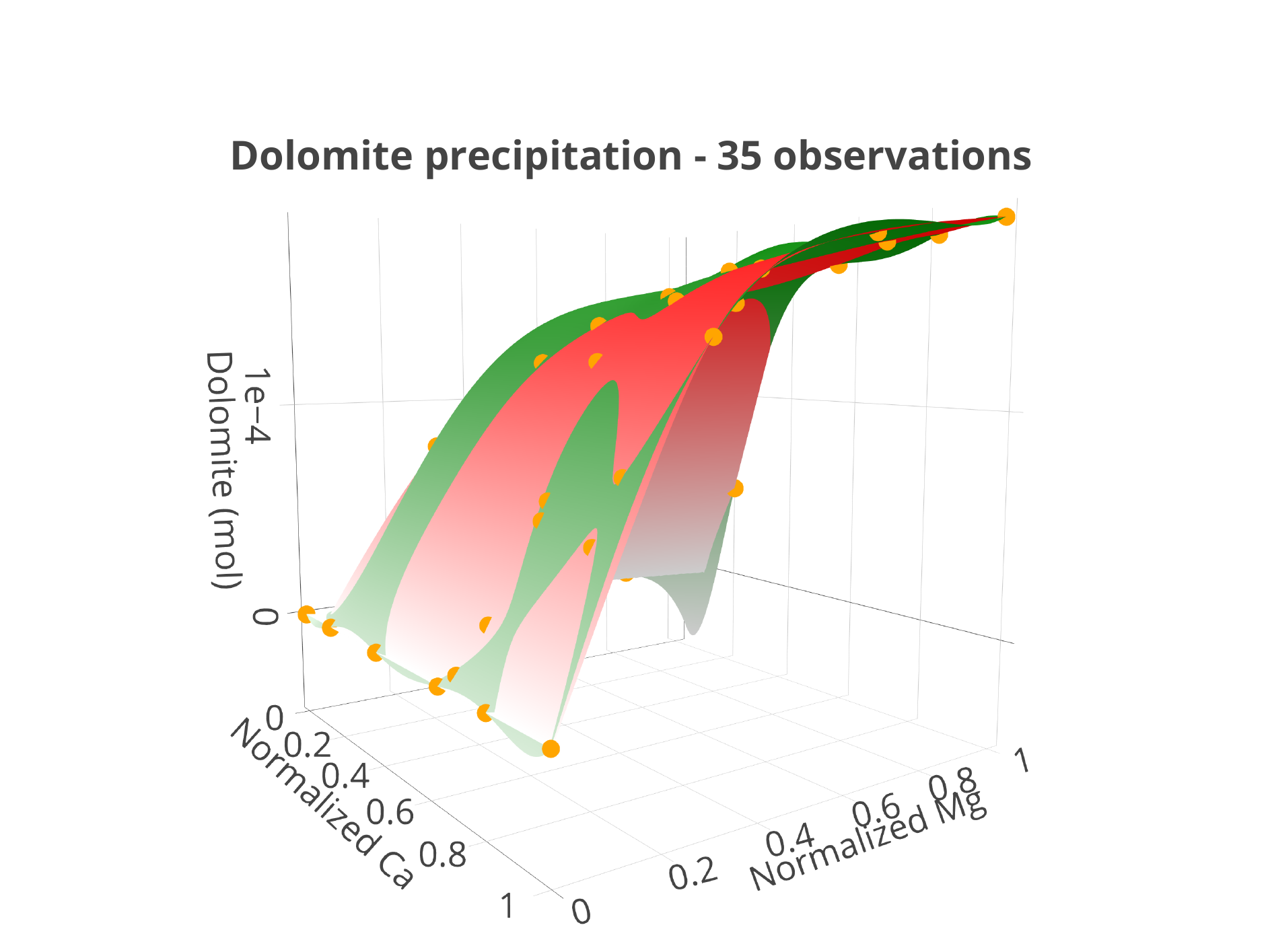} \hspace{1em}
   \includegraphics[clip, scale=0.3, trim=4cm 0.2cm 5cm 3cm]{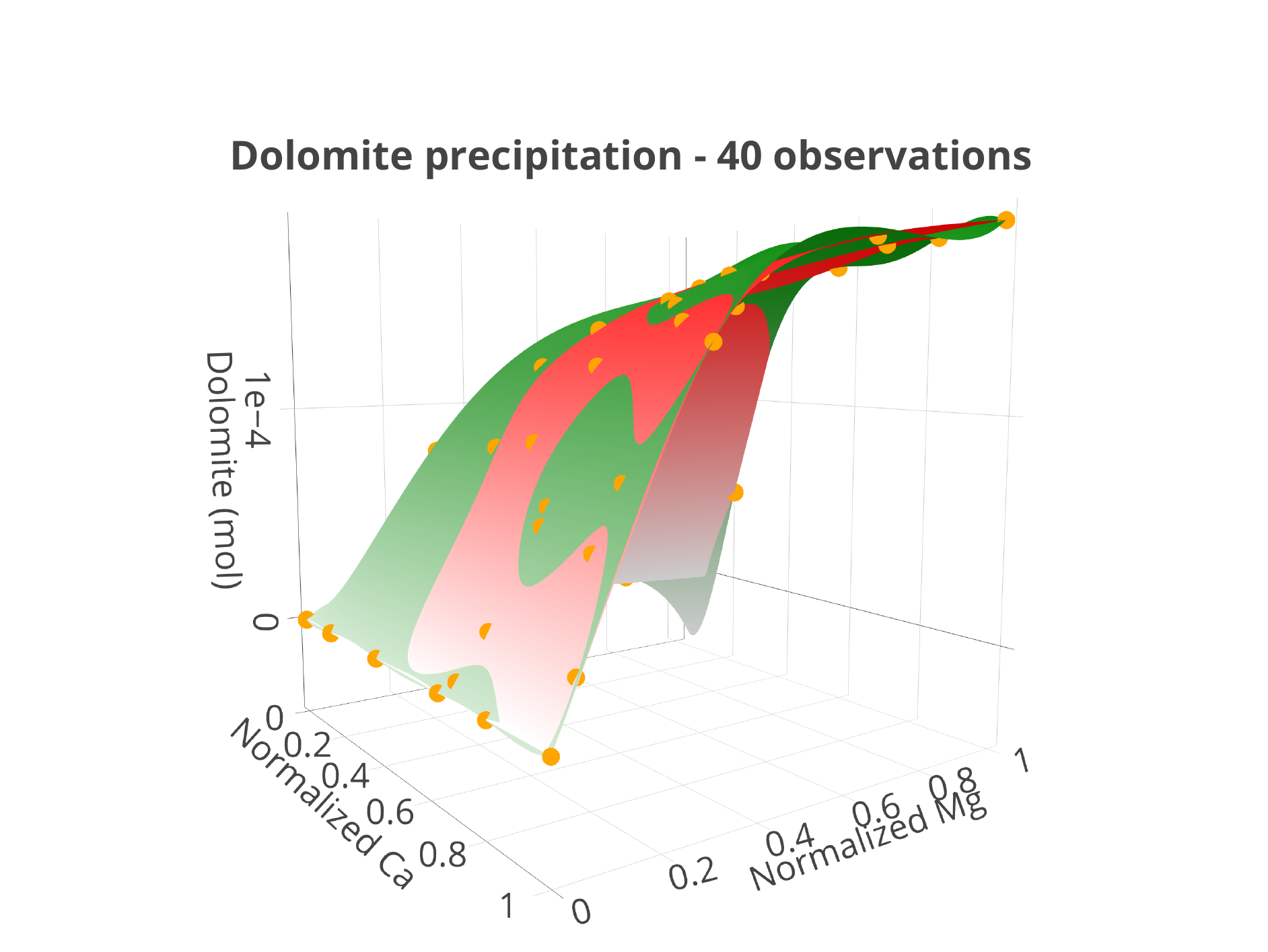}
   \includegraphics[clip, scale=0.4, trim=5cm 7.5cm 3cm 5cm]{Fig/visualisation_courbes/scale_2D.pdf} 
  \caption{Illustration of our active learning approach for estimating the function $f_2(5\times 10^{-4}, \textrm{Ca}, 2\times 10^{-3},\textrm{Mg}, 10, 0)$ displayed in the right part of Figure \ref{fig:calcite_dolomite} by starting
  from $t_1=3$ observations randomly chosen in $\textrm{A}\subset [0,1]^2$. Here, the squared exponential covariance function was used.\label{fig:plot_dolomite2D}}
\end{center}
\end{figure}
